\documentclass{article}

\usepackage[frozencache,cachedir=.]{minted}

\usepackage{microtype}
\usepackage[pdftex]{graphicx}
\usepackage{subcaption}
\usepackage{booktabs} % for professional tables

\usepackage{pifont}
\usepackage{rotating}
\usepackage{multirow}
\usepackage{customstyle}

\usepackage{pgfplots}

\pgfplotsset{compat=1.14}

\usepackage{hyperref}

\usepackage[accepted]{icml2023}

\usepackage{amsmath}
\usepackage{amssymb}
\usepackage{mathtools}
\usepackage{amsthm}
\usepackage{xspace}

\usepackage[disable,textsize=tiny]{todonotes}
\icmltitlerunning{OrionBench: Benchmarking Time Series Generative Models in the Service of the End-User}

\newcommand{\bench}{\textit{OrionBench}\xspace}

\newcommand{\user}{end-user}{}

\begin{document}

\twocolumn[
\icmltitle{OrionBench: Benchmarking Time Series Generative Models\\in the Service of the End-User}

% It is OKAY to include author information, even for blind
% submissions: the style file will automatically remove it for you
% unless you've provided the [accepted] option to the icml2023
% package.

% List of affiliations: The first argument should be a (short)
% identifier you will use later to specify author affiliations
% Academic affiliations should list Department, University, City, Region, Country
% Industry affiliations should list Company, City, Region, Country

% You can specify symbols, otherwise they are numbered in order.
% Ideally, you should not use this facility. Affiliations will be numbered
% in order of appearance and this is the preferred way.
\icmlsetsymbol{equal}{*}

\begin{icmlauthorlist}
\icmlauthor{Sarah Alnegheimish}{mit}
\icmlauthor{Laure Berti-Equille}{ird}
\icmlauthor{Kalyan Veeramachaneni}{mit}
\end{icmlauthorlist}

\icmlaffiliation{mit}{MIT LIDS, Cambridge MA}
\icmlaffiliation{ird}{IRD ESPACE-DEV, Montpellier, France}

\icmlcorrespondingauthor{Kalyan Veeramachaneni}{kalyanv@mit.edu}

% You may provide any keywords that you
% find helpful for describing your paper; these are used to populate
% the "keywords" metadata in the PDF but will not be shown in the document
\icmlkeywords{Machine Learning, ICML}

\vskip 0.3in
]

% this must go after the closing bracket ] following \twocolumn[ ...

% This command actually creates the footnote in the first column
% listing the affiliations and the copyright notice.
% The command takes one argument, which is text to display at the start of the footnote.
% The \icmlEqualContribution command is standard text for equal contribution.
% Remove it (just {}) if you do not need this facility.

\printAffiliationsAndNotice{}  % leave blank if no need to mention equal contribution
% \printAffiliationsAndNotice{\icmlEqualContribution} % otherwise use the standard text.

\begin{abstract}
Time series anomaly detection is a vital task in many domains, including patient monitoring in healthcare, forecasting in finance, and predictive maintenance in energy industries.
This has led to a proliferation of anomaly detection methods, including deep learning-based methods.
Benchmarks are essential for comparing the performances of these models as they emerge, in a fair, rigorous, and reproducible approach.
Although several benchmarks for comparing models have been proposed, these usually rely on a one-time execution over a limited set of datasets, with comparisons restricted to a few models.
We propose \bench -- an end-user centric, continuously maintained benchmarking framework for unsupervised time series anomaly detection models.
Our framework provides universal abstractions to represent models, extensibility to add new pipelines and datasets, hyperparameter standardization, pipeline verification, and frequent releases with published updates of the benchmark.
We demonstrate how to use \textit{OrionBench}, and the performance of pipelines across 17 releases published over the course of four years.
We also walk through two real scenarios we experienced with \bench that highlight the importance of continuous benchmarking for unsupervised time series anomaly detection.
\end{abstract}

\section{Introduction}

% widespread of time series anomaly detection models
As continuous data collection becomes more commonplace across domains, there is a corresponding need to monitor systems, devices, and even human health and activity in order to find patterns in collected data, as well as deviations from those patterns~\cite{chandola2009survey, aggarwal2017outlieranalysis}. Over the past decade, tremendous progress has been made in using machine learning to perform various types of monitoring, including unsupervised time series anomaly detection. 
For example, in the past 5 years, \citet{hundman2018lstmdt} created a
Long Short-Term Memory (LSTM)
forecasting model to find anomalies in spacecraft data, \citet{park2018vae} used LSTM variational autoencoders for anomaly detection in multimodal sensor signals collected from robotic arms, and \citet{geiger2020tadgan} used generative adversarial networks for time series anomaly detection on widely used public datasets.

% popularity in using unsupervised pipelines and why
These methods have gained popularity given their unsupervised nature. With supervised learning, models learn patterns from human-labelled data and then use those patterns to detect other anomalies. That these models are restricted to previously labelled events, which are themselves difficult for humans to find, makes it more challenging for models to make useful predictions. Moreover, these models struggle to find ``new'' events that are interesting to the user. In contrast, with unsupervised learning, no ground truth is given to the model, revealing anomalies that may have otherwise gone unseen.
This property is highly valuable to users, who are often unable to determine \textit{what} they are looking for and \textit{when} it will occur.
% -- which we refer to as the ``unknown-unknowns'' issue. Since unsupervised models flag any intervals that deviate from what is expected, these models can help finding ``unknown-unknowns.''
In this paper, we focus on unsupervised models.

% \begin{table*}[t]
%     \centering
%     \vspace{-1em}
%     \caption{A set of pain points experienced by users and their corresponding research questions we aim to address.}
%     \label{tab:pprq}
%     \begin{tabular}{p{2cm}p{7cm}rp{6cm}}
%     \toprule
%         & \textbf{User Pain Point}    & & \textbf{Research Question} \\
%     \midrule
%         \textbf{Abundance}      & An existing plethora of anomaly detection methods overwhelms the user. & 
%         \textbf{RQ1}            & How can we support users in finding the right model?\\
%         \textbf{Loss of Productivity}    & New methods are continually published. Users may spend substantial time trying to adapt a new model's code for their data, only to discover that the new model did not outperform their current model. & 
%         \textbf{RQ2}            & How can we provide users with pipelines that are ready off-the-shelf?\\
%         \textbf{Complexity}     & Important components (e.g. pre-processing functionalities) are hidden behind the complexity of the model, when in reality these components are what made the model successful.             & 
%         \textbf{RQ3}            & How can we best represent pipelines and attribute the successes and limitations of each component?\\
%     \bottomrule
%     \end{tabular}
% \end{table*}

\begin{table*}[t]
    \centering
    \vspace{-1em}
    \caption{A set of pain points experienced by \user{s} and their corresponding research questions we aim to address.}
    \label{tab:pprq}
    \begin{tabular}{p{1.5cm}p{8.5cm}rp{5cm}}
    \toprule
        & \textbf{End-User Pain Point}    & & \textbf{Research Question} \\
    \midrule
        \textbf{Fear of missing out}     
                                & Numerous generative modeling techniques are published, each promising better performance than all previous models. An end-user worries if the model they have been using is suboptimal and needs to updated. & 
        \textbf{RQ1}            & How can we support end-users in confidently making the decision of whether or not to adopt a new model?\vspace{3pt}\\
        \textbf{Unable to parse complex jargon}     
                                & Published work has a lot of complex machine learning specific jargon which makes it impossible for an end-user to approach implementation or delineate differences between what they have been using versus the new methods. Important components (e.g. pre-processing functionalities) are hidden behind the complexity of the model, when in reality these components are what made the model successful. \vspace{3pt}& 
        \textbf{RQ2}            & How can we best represent models with proper abstractions, such that new models can be represented as a set of components and one can identify the differences between models easily?\\
        \textbf{Time lost with no improvement}
                                & With these two challenges above, end-users may spend substantial time trying to adapt a new model's code for their data, only to discover that the new model did not outperform their current model on their data. & 
        \textbf{RQ3}            & How can we provide end-users with ready-to-use models should they choose to adopt? \\
    \bottomrule
    \end{tabular}
    \vspace{-1em}
\end{table*}

% challenges with using unsupervised models
When \textit{end-users} -- defined here as people who are interested in training a model on their own data in order to find anomalies -- attempt to use these models, they regularly run into particular challenges and pain points, which we highlight in Table~\ref{tab:pprq}. % We highlight these here. 

\begin{figure*}[t]
    \centering
    \includegraphics[width=.95\linewidth]{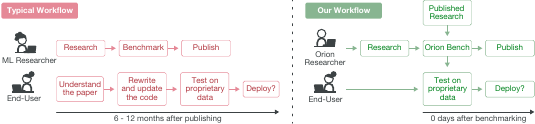}
    \caption{Typically researchers and end-users have independent processes. Researchers develop their method and benchmark it to publish their papers. Once these methods are publicized, \user{s} work on first understanding the model then adapting the code to work on their own data. After it is tested, \user{s} decide whether the performance is sufficient for it to be deployed or not. With \texttt{OrionBench}, we aim to have a single hub where researchers can benchmark their pipelines and become instantaneously available to \user{s}.}
    % \vspace{-0.5cm}
    \label{fig:workflow}
\end{figure*}

% \noindent\textbf{FOMO Challenge.} \noindent\textbf{Adoption Challenge.}
One challenge is simply deciding which model to use. Rapid innovation in the machine learning space, where papers are regularly published presenting new state-of-the-art (SOTA) models, means many users are in a constant state of struggling to keep up. Moreover, if users do decide to use the latest pipeline -- perhaps alongside their existing approach -- they often find themselves unsure of how to get started, as research papers are full of new terminology published alongside obfuscated code. 
Another challenge comes with determining whether one model works better than another. Users might find that the new model did not actually improve on their existing model. 
Lastly, and more subtly, a new model can outperform existing models not because of the model itself, but due to the inclusion of important pre- and/or post-processing operations, a distinction that may not be apparent to the user. 
% Kalyan add 
This entire process can be time-consuming, taking 6-12 months after the research is published for an \user{} to figure out and decide whether or not to incorporate a newly published method. Figure~\ref{fig:workflow}(left) depicts this asynchrony between the research process and method usage. 
% This process of an end-user trying to figure out and decide whether or not to incorporate a pipeline takes 6-12 months after the research is published. Figure 1 (Left) depicts this asynchrony between research process and usage. 

It is important to note that an \user{} is only interested in finding the best solution for their particular problem, and does not think about these modeling techniques in the way that researchers do. In recent years, benchmarks have become instrumental in gauging and comparing model performance for machine learning researchers ~\cite{coleman2019dawnbench, han2022adbench}. In this paper, we ask — can benchmarking frameworks also help to alleviate \user{s} challenges? 
%Kalyan add after can benchmarking alleviate...?
Specifically, we ask if it possible to bring together both the researcher and the \user{} to utilize benchmarking systems and reduce the time required to use the latest research. 
% We attempt to achieve this through our benchmarking workflow shown in Figure~\ref{fig:workflow} on the right. 

% In this paper, 
We propose \bench -- an end-user centric benchmarking framework for unsupervised time series anomaly detection. 
% To develop a system that addresses our research questions in Table~\ref{tab:pprq}, we investigate the current workflow and objective of benchmarks.
Figure~\ref{fig:workflow}(left) illustrates how in a regular setting researchers and \user{s} operate independently from one another, creating hurdles for \user{s} when adopting a new published model. 
With our system Figure~\ref{fig:workflow}(right), all researchers (Orion and ML researchers) directly contribute their model to \textit{OrionBench}. This creates a single source of readily-available models for the \user{}.
% Still, we propose that with a few careful changes, these benchmarking frameworks can further help \user{s}. 
% With that in mind, 

Three concrete innovations enable us to address \user{s’} common but critical concerns: \vspace{-1em}
\begin{itemize}
\itemsep0em
\item A continuously running system, moving away from \textit{point-in-time} evaluations.  
\item Abstractions that allow us to easily incorporate and assess new models, and isolate the factors that make them better.
\item Seamless integration of the latest models into usable \textit{pipelines} for end-users. 
\end{itemize}

Below we highlight our framework's unique contributions. \bench is: 
\begin{enumerate}
    \item \textbf{A standardized framework} that enables the integration of new pipelines and datasets. \bench started with 2 pipelines in 2020, and as of now encompasses 12 pipelines, 28 primitives, 14 public datasets, and 2 custom evaluation metrics. 
    Once integrated into the framework and benchmarked, a pipeline is seamlessly made available to the \user{} through unified APIs.
    \item \textbf{A continuously-run benchmark} with frequent releases. To date, we have 17 benchmark leaderboards covering almost four years, accumulating over 70,032 experiments.
    % starting in September 2020. Our latest release in October 2024 makes for an overall . 
    In addtion, we demonstrate the stability and reproducibility of \bench. 
    % by showcasing the progression of pipeline performances.
    % across all releases to date.
    \item \textbf{An end-to-end benchmark} executable with a single command. 
    Given a pipeline and datasets, the benchmark evaluates the performance of the pipeline on every signal according to time series anomaly detection-based metrics. 
    We provide an extensive evaluation that illustrates the qualitative and computational performances of pipelines across all datasets according to time series anomaly detection-based metrics.
    \item \textbf{Open-source} and publicly available: \url{https://github.com/sintel-dev/Orion}.~\footnote{Reproducing paper figures and tables is available: \url{https://github.com/sarahmish/orionbench-paper}}
\end{enumerate}

% section description
% The remainder of the paper is organized as follows:
% In Section~\ref{sec:related}, we compare \bench to existing open-source benchmarking frameworks. We detail the user-centric components of \bench in Section~\ref{sec:system}. Section~\ref{sec:demo} presents an evaluation of our benchmarking framework and describes two real scenarios in which \bench was valuable. We conclude in Section~\ref{sec:discussion}.

\section{\bench}
\label{sec:system}

\begin{figure}[t]
    \centering
    \includegraphics[width=.8\linewidth]{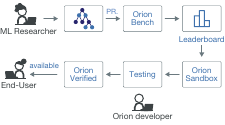}
    \caption{OrionBench integrates new models made by ML researchers and compares its performance to currently available models through the leaderboard. After testing the validity and reproduciblity of the model, it is transferred from ``sandbox'' to ``verified'' and becomes readily available to the end-user.}
    \label{fig:userbench}
    % \vspace{-2em}
\end{figure}

\bench is a benchmark suite within the Orion system~\cite{alnegheimish2022orion}.
A researcher creates a new model and integrates it with Orion through a pull request.
A benchmark run is executed and produces a leaderboard, and the model is then stored in the sandbox.
This part of the workflow satisfies the goals of the researcher, which is comparing the performance of different models.
To serve \user{s}, pipelines in the sandbox are tested by an Orion developer. 
Pipelines that pass the tests are verified and become available to \user{s}.
This workflow is depicted in Figure~\ref{fig:userbench}.
% With a single command line, users are able to run the benchmark end-to-end.
Five main properties enable our framework for benchmarking unsupervised time series anomaly detection models:
abstractions that enable us to compose models as pipelines (directed acyclic graphs) of reusable components called \textit{primitives}; hyperparameter standardization; extensions to add new pipelines and datasets; verification of pipelines; and continuous benchmark releases. Lastly, we conclude the section by illustrating how \bench benefits the \user{}.

% \vspace{-3em}
\subsection{Abstracting Models into Primitives and Pipelines}
\label{sec:abstractions}
% how do we represent models? though primitives and pipelines.

New unsupervised time series anomaly detection models are constantly being developed. 
This poses the question: \textit{how do we uniformly represent these models?}

To accomplish this goal, we standardize models.
The anomaly detection process starts with a signal $\mathbf{X} = \{\mathbf{x}_1, \mathbf{x}_2, \dots, \mathbf{x}_T\}$ where $T$ is the length of the time series and $\mathbf{x}_i \in \mathbb{R}^n$ and $n$ is the number of channels.
When the time series is a univariate signal then $n = 1$.
The goal is to find a set of anomalous intervals $A = \{(t_s^1, t_e^1), \dots, (t_s^k, t_e^k)\}$ where $k \geq 0$.
Each interval represents the start and end \textit{timestamps} of the detected anomaly.

\begin{figure*}
\centering
% \vspace{-2em}
\begin{subfigure}{.28\linewidth}
  \centering
  \includegraphics[width=.60\linewidth]{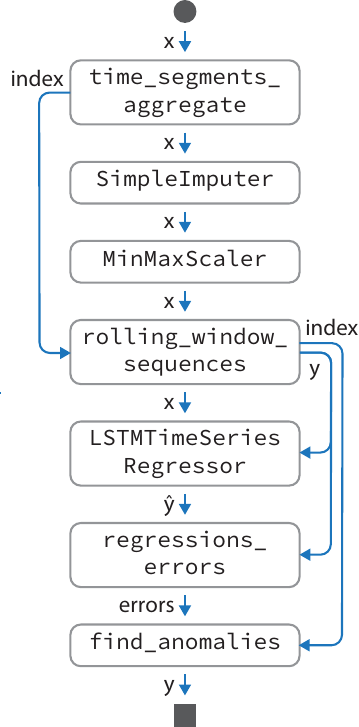}
  \caption{Graph representation}
  \label{fig:pipeline}
\end{subfigure}%
\hfill
\begin{subfigure}{.28\linewidth}
  \centering
\begin{minted}{python}
from orion import Orion
from orion.data import load_signal

train = load_signal('S-1-train')
test = load_signal('S-1-test')

orion = Orion(
    pipeline='lstm_dt'    
)

# train pipeline
orion.fit(train)

# detect anomalies
anomalies = orion.detect(test)
\end{minted}
  \vspace{1em}
  \caption{Usage in \texttt{python}}
  \label{fig:orion_api}
\end{subfigure}%
\hfill
\begin{subfigure}{.28\linewidth}
  \centering
\begin{minted}{python}
hyperparameters = {
    'time_segments_aggregate': {
        'interval': 21600
    },
    'rolling_window_sequences': {
        'target_column': 0, 
        'window_size': 250
    },
    'LSTMTimeSeriesRegressor': {
        'epochs': 35
    },
    'find_anomalies': {
        'window_size_perc': 30, 
        'fixed_threshold': false
    }
}
\end{minted}
  \vspace{1em}
  \caption{Hyperparameter config}
  \label{fig:hyper_config}
\end{subfigure}
\caption{Example of \texttt{LSMT DT} pipeline. (a) Graph representation of the pipeline showcasing its primitives and data flow. (b) \texttt{python} usage example. (c) Subset of hyperparameter configuration in \texttt{json} format of the pipeline.}
% \vspace{-1em}
\end{figure*}

We adopt a universal representation of \textit{primitives} and \textit{pipelines}~\cite{smith2020mlbazaar}.
\textit{Primitives} are reusable basic block components that perform a single operation. 
Primitives can be single tasks, and range from data scaling to signal processing to model training.
When primitives are stacked together, they compose \textit{pipelines}.
A pipeline is computed into its respective computational graph, similar to the \texttt{LSTM DT} pipeline and its primitives shown in Figure~\ref{fig:pipeline}, 
where the input is a uni- or multi-variate time series, and the output is a list of intervals of the detected anomalies.
As portrayed in Figure~\ref{fig:orion_api}, we use the \texttt{fit} method to train the model and the \texttt{detect} method to run inference.
With this standardization, we are able to treat all models equivalently.
Primitives provide a code-efficient structure such that we can be modular and re-use primitives between pipelines.
Moreover, it allows researchers to conduct ablation studies in order to attribute pipeline performance and the contribution of primitives.

\subsection{Standardizing Hyperparameter Settings}
\label{sec:hyper_stand}
Deep learning models require setting a multitude of hyperparameters, some of which are model-specific.
% Moreover, each model contains hyperparameters that are model-specific.
This has made it more challenging to keep benchmarks fair and transparent.
In \bench, hyperparameters are stored as \texttt{json} files to expose configurations in both machine- and human- readable representations.
% In \bench, hyperparameters are stored as \texttt{json} files with their configuration settings.
% With a \texttt{json} format, hyperparameters are exposed in both machine- and human- readable representations.
Figure~\ref{fig:hyper_config} is an example of the hyperparameter settings for \texttt{LSTM DT}.

To increase benchmark fairness, we standardize hyperparameters for both \textit{global} and \textit{local} hyperparameters.
\textbf{Global} hyperparameters are shared between pipelines. They typically pertain to pre- and post- processing primitives.
For example, in Figure~\ref{fig:hyper_config}, \texttt{interval} is a global hyperparameter that denotes the aggregation level for the signal -- here it is set to 6 hours of aggregation (21,600 seconds).
Such hyperparameters are selected based on the characteristics of the dataset, and in some cases are dynamic.
For example, \texttt{window\_size\_perc} sets the window size to be 30\% of the signal length.
\textbf{Local} hyperparameters such as \texttt{epochs} are pipeline-specific and are selected based on the authors' recommendation in the original paper.
These hyperparameters are consistent across datasets per pipeline in order to alleviate any bias introduced by knowing the ground truth anomalies of the dataset.

% OrionBench is specifically made for \textit{unsupervised} time series anomaly detection. 
% As suggested in~\cite{alnegheimish2022sintel}, hyperparameter tuning can be introduced in unsupervised learning by improving the underlying model (e.g. \texttt{LSTMTimeSeriesRegressor}) to forecast a signal that highly resembles the original one by tuning the hyperparameters to minimize metrics such as mean squared error.
% However, such procedures can overfit the signals to anomalies which precludes subsequent primitives from finding these anomalies, impairing the pipeline's performance.
% As such, hyperparameter tuning is not currently introduced within the benchmark.

\subsection{Integrating New Pipelines and Datasets}
A main pillar of open-source development is continuously maintaining and updating a library.
Benchmark libraries are no different.
For a library to grow, it is essential to keep introducing new pipelines and datasets to benefit the end-user.

ML researchers build new primitives and compose new pipelines easily in \bench.
The framework provides templates to help guide researchers in this process.
Moreover, ML researchers can utilize primitives in other packages given a corresponding \texttt{json} representation.
% They can also use readily available primitives within their pipelines~\cite{smith2020mlbazaar}.
It is often the case that pre- and post- processing primitives are reusable across pipelines~\cite{alnegheimish2022sintel}.
\bench first started with 2 pipelines, and now has 12 pipelines.
The same applies to benchmark datasets.
To make the data more accessible, we host publicly available datasets on an Amazon S3 instance.
Signals can be loaded via \texttt{load\_signal} command (as shown in Figure~\ref{fig:orion_api}) that will directly connect to S3 if the data is hosted there.
Otherwise, it will search for the file locally.
This enables users to also load their own private custom data for benchmarks.

%note for Sarah: For future this section can be improved a lot. I think one common problem we notice is that when folks run benchmarking, they sometimes manually change the hyperparameters in their desktop, but never check those values in. And this across users the results differ. I think the standardization helps with that as well. 

\subsection{Verifying Pipelines}
We organize pipelines into \textit{verified} pipelines and \textit{sandbox} pipelines.
When a new pipeline is proposed, it is categorized under ``sandbox" until several tests and validations are made. 
The ML researcher opens a new pull request and is requested to pass unit and integration tests before the pipeline is merged and stored in the sandbox. 
Next, Orion developers test the new pipeline and verify its performance and reproducibility. One of the most commonly encountered situations is a mismatch between the researchers' comparison report and the results an Orion developer would get from running the same framework. A very common reason for this was
that researchers had failed to update a hyperparameter setting. 
Once these checks are made, pipelines are transferred from ``sandbox'' to ``verified''. 
The increased reliability of verified pipelines enhances the end-user's confidence in adopting pipelines.

\subsection{Releasing Regularly}
\label{sec:releases}
The last requirement for an end-user-friendly benchmarking framework like \bench is to keep track of how benchmark results change over time.
Most pipelines are stochastic in nature, meaning benchmark results can change from run to run.
Moreover, when the underlying dependency packages (e.g. \texttt{TensorFlow}) introduce new versions, benchmark results can be affected or even compromised.
Therefore, it is crucial to monitor the pipeline performances over time and prevent possible breakdowns due to backwards incompatibility.

This need is a main driver behind the creation of \textit{OrionBench}.
Benchmarking was introduced as a measure of \textit{stability} and \textit{reproducibility} testing, analogous to how Continuous Integration Continuous Deployment (CI/CD) tests have greatly increased the reliability of open-source libraries.
\bench now serves as a test of pipeline stability over time.
As of now, 17 releases have been published, and the \textit{leaderboard} changes with each release (see Section~\ref{sec:stability}).

\subsection{Benefiting the End-User}
\label{sec:benefits}

\bench is available to the end-user on pypi, where they can install \bench through ``pip install Orion''. Then all verified pipelines are at the finger tips of the end-user where they train a pipeline and detect anomalies using \texttt{fit} and \texttt{detect} APIs, respectively. End-Users have access to a collection of models that they trust to perform as expected, fits their computational needs, and are continuously maintained and benchmarked.
\section{Evaluation}
\label{sec:demo}

We demonstrate the use of \bench on 12 pipelines ranging from classic to generative models and 14 datasets.
We also lay out how benchmarking works as a mechanism to test pipeline stability.
Moreover, we present two real-world scenarios in which \bench was used to ground unsupervised anomaly detection.

\renewcommand*{\thefootnote}{\arabic{footnote}}

\begin{table}[t]
    % \small
    \centering
    \caption{Datasets Summary. There are 14 datasets with varying number of signals and anomalies. The table presents the average signal length and anomaly length for each dataset. All these datasets are publicly accessible.}
    \label{tab:datasets}
    \resizebox{\linewidth}{!}{%
    \begin{tabular}{llccc}
    \toprule
                                    & Dataset                   & \# Signals   & \# Anomalies   & Avg. Signal           \\
    \midrule
         \multirow{2}{*}{NASA}      & MSL                       & 27        & 36                & 4890.59               \\
                                    & SMAP                      & 53        & 67                & 10618.86              \\
    \midrule
         \multirow{5}{*}{NAB}       & Art                       & 6         & 6                 & 4032.00               \\
                                    & AWS                       & 17        & 30                & 3980.35               \\
                                    & AdEx                      & 5         & 11                & 1593.40               \\
                                    & Traf                      & 7         & 14                & 2237.71               \\
                                    & Tweets                    & 10        & 33                & 15863.1               \\
    \midrule
         \multirow{4}{*}{Yahoo S5}  & A1                        & 67        & 178               & 1415.9                \\
                                    & A2                        & 100       & 200               & 1421.0                \\
                                    & A3                        & 100       & 939               & 1680.0                \\
                                    & A4                        & 100       & 835               & 1680.0                \\
    \midrule
        \multirow{3}{*}{UCR}        & Natural                   & 142       & 142               & 99973.33              \\
                                    & Distorted                 & 92        & 92                & 49218.02              \\
                                    & Noise                     & 16        & 16                & 39343.38              \\
    \midrule
        Total                       &                           & 742       & 2599              &\\
    \bottomrule
    \end{tabular}}
    \vspace{-1.5em}
\end{table}

\textbf{Datasets.}
Currently, the benchmark is executed on 14 datasets with ground truth anomalies.
These datasets are gathered from different sources, including 
\textbf{NASA}~\footnote{\url{https://github.com/khundman/telemanom}}, 
\textbf{NAB}~\footnote{\url{https://github.com/numenta/NAB}}, \textbf{UCR}~\footnote{\url{https://www.cs.ucr.edu/~eamonn/time_series_data_2018}},
and \textbf{Yahoo S5}~\footnote{\url{https://webscope.sandbox.yahoo.com/catalog.php?datatype=s&did=70}}.
Collectively, these datasets contain 742 time series and 2,599 anomalies.
The properties of each dataset, including the number of signals and anomalies, the average length of signal, and the average length of anomalies are presented in Table~\ref{tab:datasets}.
The table makes clear how properties differ between datasets; for instance, NASA, NAB, and UCR contain anomalies that are longer than those in Yahoo S5, and the majority of anomalies in Yahoo S5's A3 \& A4 datasets are point anomalies.
% It is clear to see that Yahoo S5 contains different types of anomalies compare to NASA and NAB.

\textbf{Models.}
As of the writing of this paper, \bench includes 12 pipelines:
% \textbf{P}rediction, \textbf{R}econstruction-, \textbf{D}ensity-, and \textbf{S}ervice-based models:
\texttt{ARIMA} -- 
Autoregressive Integrated Moving Average statistical model~\cite{box1968some};
\texttt{MP} -- Discord discovery through Matrix Profiling~\cite{yeh2016matrixprofile};
\texttt{AER} --
AutoEncoder with Regression deep learning model with reconstruction and prediction errors~\cite{wong2022aer};
\texttt{LSTM-DT} --
LSTM non-parametric Dynamic Threshold with two LSTM layers~\cite{hundman2018lstmdt};
\texttt{TadGAN} --
Time series Anomaly Detection using Generative Adversarial Networks~\cite{geiger2020tadgan};
\texttt{LSTM VAE} --
Variational AutoEncoder with LSTM layers~\cite{park2018vae};
\texttt{LSTM AE} --
AutoEncoder with LSTM layers~\cite{malhotra2016lstmae};
\texttt{Dense AE} --
Similar to \texttt{LSTM AE}, with Dense layers~\cite{sakurada2014dense};
\texttt{LNN} --
Liquid Neural Network model, a variant of Liquid Time-Constant Networks~\cite{hasani2021lnn};
\texttt{GANF} --
Graph Augmented Normalizing Flows density-based model~\cite{dai2022graph};
\texttt{AT} -- AnomalyTransformer model with association discrepancy~\cite{xu2022anomalytransformer};
\texttt{Azure AD} --
Microsoft Azure Anomaly Detection service~\cite{ren2019azure}.
% These pipelines were introduced at different stages of development. Section~\ref{sec:stability} provides further details about when exactly each model was integrated.

\begin{figure*}[t]
    \centering
    \includegraphics[width=\textwidth]{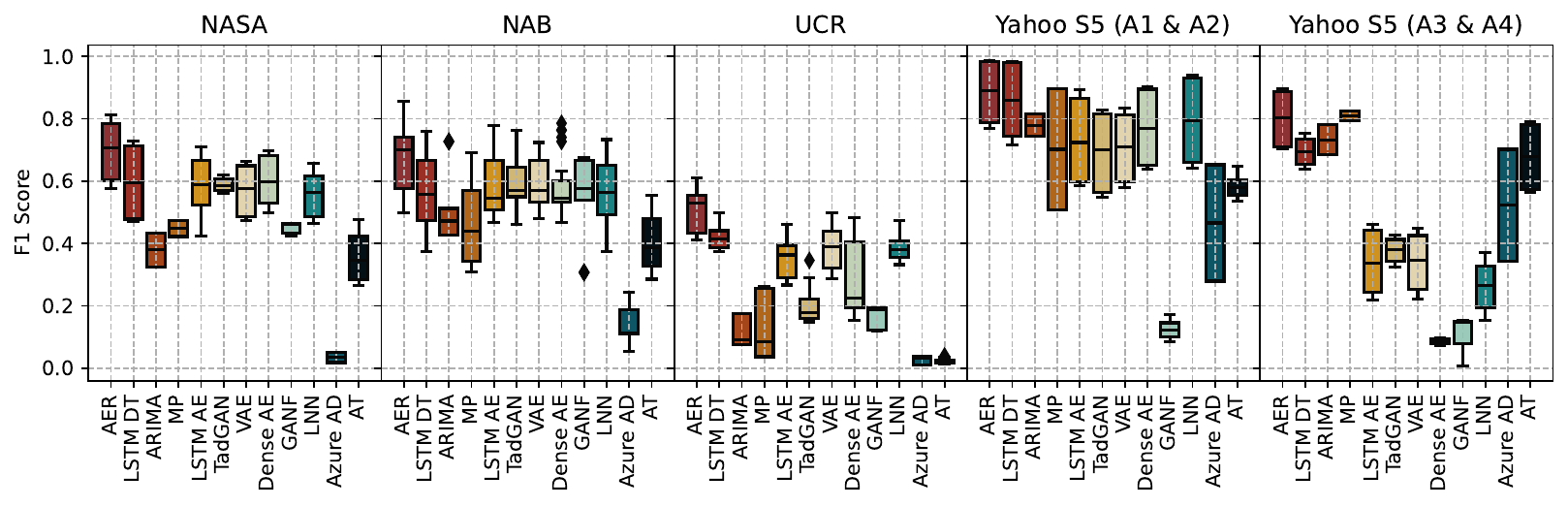}
    \caption{Distribution of F1 Scores across NASA, NAB, Yahoo S5, and UCR. Yahoo S5 was split into two subsets highlighting the F1 difference pipelines experience when detecting point anomalies.}
    \label{fig:f1score}
    \vspace{-1em}
\end{figure*}

\textbf{Hyperparameters.}
Hyperparameter settings are an important part of model performance.
As highlighted in Section~\ref{sec:hyper_stand}, \bench seeks to provide a fair benchmark by standardizing hyperparameters.
Moreover, we sort pipelines based on whether they are prediction-based or reconstruction-based~\cite{alnegheimish2022sintel} and set the hyperparameters based on those properties.
The precise values are selected based on the configurations proposed by the original authors in previous work~\cite{hundman2018lstmdt, wong2022aer, geiger2020tadgan, dai2022graph}.
For example, reconstruction-based pipelines tend to have a smaller \texttt{window\_size} compared to prediction-based pipelines given that reconstructing large segments is difficult.
For example, prediction-based pipelines have a \texttt{window\_size} of 250 data points, while reconstruction-based pipelines have a smaller \texttt{window\_size} of 100, because the objective is to reconstruct the entire window rather than predict a few steps ahead.
While some other methods alter the \texttt{window\_size} based on the signal length~\cite{malhotra2016lstmae}, we provide the option to make these hyperparameters dynamic.
For example, \texttt{window\_size} can be set as 10\% of the entire signal length.

\textbf{Settings.} We use \bench version \texttt{0.5.2}. The benchmark is executed to run for 5 iterations over all the pipelines and datasets.

\textbf{Compute.}. We set up an instance on the MIT SuperCloud~\cite{reuther2018supercloud} with an Intel Xeon Gold 6249 processor of 10 CPU cores (9 GB RAM per core) and one NVIDIA Volta V100 GPU.

\begin{figure}[t]
\begin{minted}{python}
from orion.benchmark import benchmark

pipelines = [ 'arima', 'lstm_dt', 'aer']
datasets = ['NAB', 'NASA']
metrics = ['f1', 'precision']

benchmark(pipelines=pipelines, datasets=datasets, metrics=metrics, rank='f1')
\end{minted}
\vspace{-1em}
\caption{Benchmark command in \texttt{Python}. Running \texttt{benchmark()} with default settings will execute the benchmark on all pipelines and datasets currently integrated.}
\vspace{-1em}
\label{fig:benchmark_api}
\end{figure}

\subsection{End-to-End Benchmark}
\label{sec:benchmark}
The script in Figure~\ref{fig:benchmark_api} illustrates the few lines of code required for execution.

\textbf{Benchmark Usage.} \bench is available to all users through a single command, as illustrated in Figure~\ref{fig:benchmark_api}.
Users specify the list of pipelines, datasets, and metrics they are interested in and pass them to the \texttt{benchmark} function.
The output result is stored as a detailed \texttt{.csv} file that signals performance metrics for each pipeline, such as accuracy, precision, recall, and F1 score.
It also shows the status of the run -- whether it was successful or not, total execution time, and the runtime for each internal primitive.

\textbf{Qualitative Performance.} 
% the performance of pipelines is sensetive to the dataset.
% AER is the best pipeline across all datasets on average.
% there are pipelines that really struggle with point anomalies like LSTM AE, TadGAN, and VAE which are all reconstruction based methods with internal LSTM layers.
% Azure is working for yahoo because it frequently flags intervals as anomalous.
% Once we obtain the detailed sheet, we can aggregate the results per pipeline and dataset.
% Typically, we are interested in the F1 score; however, users can alter the metric of interest.
% \bench summarizes the results automatically to produce: (1) an overview of the performance of each pipeline on each dataset, and (2) a leaderboard showing the number of times a pipeline surpassed \texttt{ARIMA} in each one of the benchmark datasets.
% The final step of the benchmark produces a detailed \texttt{.csv} file that is later summarized automatically.
Figure~\ref{fig:f1score} depicts the F1 score obtained for each dataset on average.
The score achieved by each pipeline differs based on the dataset and its properties.
We can see that \texttt{AER} is the highest-performing pipeline overall. 
% although \texttt{LSTM DT} competes with \texttt{AER} in some cases.
Another interesting observation is that \texttt{LSTM AE}, \texttt{TadGAN}, \texttt{VAE}, and \texttt{Dense AE} are not effective at detecting point anomalies.
These pipelines are all reconstruction-based and are susceptible to anomalous regions when computing the deviation between the original and reconstructed signal, producing anomaly scores with reduced peaks at these points. Anomalies thus pass by undetected~\cite{wong2022aer}.
This is clearly demonstrated in the Yahoo S5 datasets, where F1 scores for A3 \& A4 datasets are low compared to those for A1 \& A2.
Furthermore, the \texttt{Azure AD} pipeline frequently flags segments as anomalous. 
This strategy works for datasets with a lot of anomalies, such as Yahoo S5. 
We therefore notice an increased F1 score there compared to other datasets.

% \begin{table}[ht]
%     \centering
%     \caption{Leaderboard showing number of datasets in which each pipeline outperformed~\texttt{ARIMA}.}
%     \label{tab:leaderboard}
%     \begin{tabular}{rc}
%         \toprule                                            
%                                                             &   Outperform \\
%         \cmidrule(lr){2-2}
%         \textbf{Pipeline}                                   &   \texttt{ARIMA, 1970}~\cite{box1968some}  \\
%         \midrule
%         \texttt{AER, 2022}~\cite{wong2022aer}               & 13 \\
%         \texttt{LSTM DT, 2018}~\cite{hundman2018lstmdt}     & 10 \\
%         \texttt{LSTM AE, 2016}~\cite{malhotra2016lstmae}    & 9 \\
%         \texttt{TadGAN, 2020}~\cite{geiger2020tadgan}       & 9 \\
%         \texttt{VAE, 2018}~\cite{park2018vae}               & 9 \\
%         \texttt{Dense AE, 2014}~\cite{sakurada2014dense}    & 9 \\
%         \texttt{LNN, 2021}~\cite{hasani2021lnn}             & 9 \\
%         \texttt{GANF, 2022}~\cite{dai2022graph}             & 8 \\
%         \texttt{MP, 2016}~\cite{yeh2016matrixprofile}       & 7 \\
%         \texttt{AT, 2022}~\cite{xu2022anomalytransformer}   & 2 \\
%         \texttt{Azure AD, 2019}~\cite{ren2019azure}         & 0 \\
%         \bottomrule
%     \end{tabular}
% \end{table}
\begin{table}[t]
    \small
    \centering
    \caption{Leaderboard showing number of datasets in which each pipeline outperformed~\texttt{ARIMA}.}
    \label{tab:leaderboard}
    \begin{tabular}{rc}
        \toprule                                            
                                                            &   Outperform \\
        \cmidrule(lr){2-2}
        \textbf{Pipeline}                                   &   \texttt{ARIMA, 1970}\\&~\cite{box1968some}  \\
        \midrule
        \texttt{AER, 2022}          & 13 \\
        \texttt{LSTM DT, 2018}      & 10 \\
        \texttt{LSTM AE, 2016}      & 9 \\
        \texttt{TadGAN, 2020}       & 9 \\
        \texttt{VAE, 2018}          & 9 \\
        \texttt{Dense AE, 2014}     & 9 \\
        \texttt{LNN, 2021}          & 9 \\
        \texttt{GANF, 2022}         & 8 \\
        \texttt{MP, 2016}           & 7 \\
        \texttt{AT, 2022}           & 2 \\
        \texttt{Azure AD, 2019}     & 0 \\
        \bottomrule
    \end{tabular}
    \vspace{-2em}
\end{table}

\textbf{Leaderboard.}
Table~\ref{tab:leaderboard} compares the performance of each pipeline and reports the number of datasets for which it outperforms \texttt{ARIMA}.
Out of 14 datasets, \texttt{AER} is outperforming \texttt{ARIMA} for 13, while \texttt{Azure AD} is performing worse for all datasets.
Since the benchmark was executed for 5 iterations, we report the median result.
However, pipelines are stochastic and might perform differently between runs.
Therefore, \user{s} are left to wonder whether the rankings provided in the leaderboard are robust and trustworthy.
Using Spearman's rank correlation, $\rho = 0.916$, we find that the best pipelines are consistent across runs. % (min $\rho = 0.845$,  max $\rho = 0.972$).
Similarly, pipelines at the lower end of the table are stable in their rankings.
On the other hand, the middle part of the table is subject to change as \texttt{TadGAN}, \texttt{LSTM AE}, \texttt{VAE}, \texttt{LNN}, and \texttt{Dense AE} compete with one another. 
The exact ranking of each pipeline in all runs is shown in Table~\ref{tab:ranks} in the Appendix.

\begin{figure}[t]
    \centering
    \includegraphics[width=.95\linewidth]{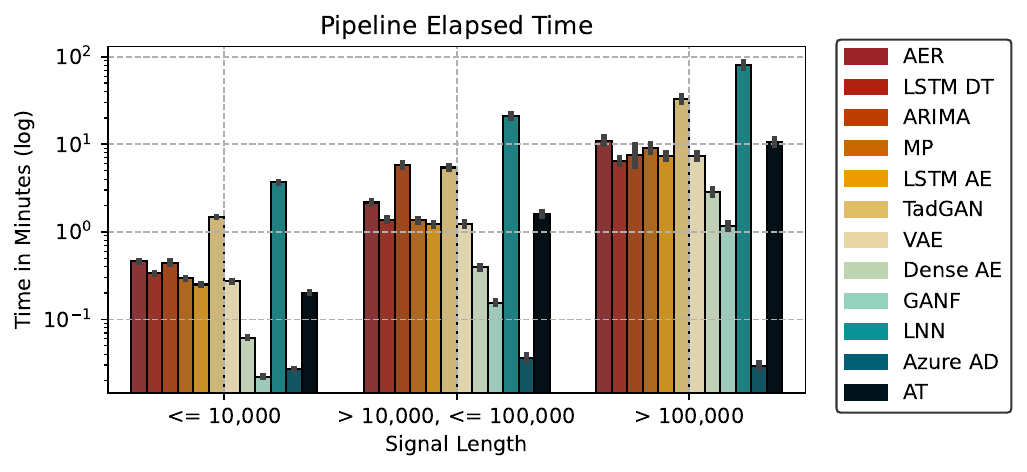}
    % \vspace{-1em}
    \caption{Average elapsed time of pipelines across different signal length groups.\vspace{-2em}}
    \label{fig:elapsed}
\end{figure}

\textbf{Computational Performance.} 
% A trade-off between runtime and performance.
In addition to quality performance, \user{s} are interested in pipelines' computational performance.
Figure~\ref{fig:elapsed} illustrates how much time (in minutes) on average each pipeline needs depending on the signal length.
Elapsed time includes the time it takes to train a pipeline and time it takes to run inference.
The shortest signal in all datasets contains 750 data points, while the longest one contains 900,000 data points. 
On shorter signals, pipelines typically take seconds, while longer signals may take minutes or even hours to complete.
The most time-consuming pipeline is \texttt{LNN} and in second place is \texttt{TadGAN}, which has more neural networks to train than other pipelines. 
% (an encoder, a generator, and 2 critics).
On the longest signal in the dataset, it takes \texttt{LNN} and \texttt{TadGAN} approximately 5 and 2 hours total elapsed time respectively.
Moreover, inference-only pipelines, such as \texttt{Azure AD}, are computationally fast and almost invariant to the length of the signal.
some pipelines can become more demanding when the signal length increases such as \texttt{AT} where the runtime increased by a factor of 7$\times$ and 8$\times$ respectively.

Given the complexity of this model, \user{s} might want to select an alternative pipeline.
Moreover, a user might sacrifice quality performance for computational efficiency, or vice versa. Individual end users can make their own decisions when weighing these tradeoffs.

% \begin{figure}[t]
% \centering
% \begin{subfigure}{.5\linewidth}
%   \centering
%   \includegraphics[width=\linewidth]{figures/versionsub.pdf}
%   \vspace{0em}
%   \caption{}
%   \label{fig:version}
% \end{subfigure}%
% \begin{subfigure}{.5\linewidth}
%   \centering
%   \includegraphics[width=\linewidth]{figures/change.pdf}
%   \caption{}
%   \label{fig:change}
% \end{subfigure}%
% \caption{Monitoring pipelines' performance across releases. (a) Average F1 scores of \texttt{AER}, \texttt{LSTM DT}, \texttt{ARIMA}, and \texttt{Azure} across multiple releases. (b) Difference in F1 score between releases.}
% \end{figure}

\subsection{Progression of Benchmarks}
\label{sec:stability}

% biggest change is [0.1.3 to 0.1.4 due to change in calculating overall performance metrics] and [0.5.0 to 0.5.1 due to adding a new dataset that is considerably harder the the previous ones.]
% there are fluctuations but they are not severe with a mean around 0.

\textbf{Stability.} As we see in Figure~\ref{fig:f1score}, \texttt{AER} is the highest-performing pipeline: \textit{Was this always the case?} 
\bench publishes benchmark results with every package release.

\begin{figure}[t]
    \centering
    \includegraphics[width=0.83\linewidth]{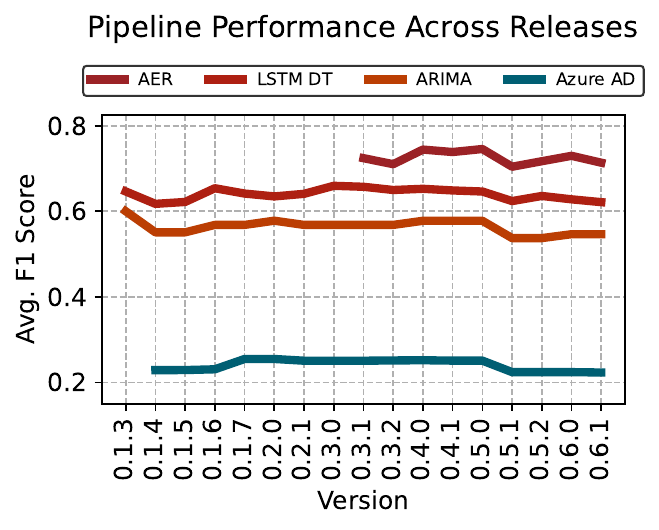}\vspace{-1em}
    \caption{Monitoring pipelines' performance across all formal releases of \bench.}
    \label{fig:version}
\end{figure}

Figure~\ref{fig:version} depicts the average F1 score of four pipelines.
These pipelines are chosen to show: the best-performing pipeline (\texttt{AER}), the worst-performing pipeline (\texttt{Azure AD}), the first implemented pipeline (\texttt{LSTM DT}), and the classic pipeline (\texttt{ARIMA}), that are currently available in our framework.
The observed performance can change from one release to another for a number of reasons, including the stochastic nature of pipelines, internal changes in dependency packages, and dynamic thresholding.
If we look closely at Figure~\ref{fig:version}, we notice three shifts (viewed as slopes) to \texttt{LSTM DT}, and only two to \texttt{AER}, \texttt{ARIMA}, and \texttt{Azure AD}.

\begin{figure*}[t]
    \centering
    \small
    \begin{tikzpicture}
\usetikzlibrary{calc}

% draw arrow
\coordinate (start) at (-11,0);
\coordinate (end) at (7,0);
\draw [line width=1pt, -stealth, gray] (start) -- (end);

% You can use `foreach` to improve the following codes
\coordinate (s0) at (-10,0);
\coordinate (t0) at ($(s0)+(0,0.2)$);
\coordinate (s1) at (-8,0);
\coordinate (t1) at ($(s1)+(0,0.2)$);
\coordinate (s2) at (-6,0);
\coordinate (t2) at ($(s2)+(0,0.2)$);
\coordinate (s3) at (-4,0);
\coordinate (t3) at ($(s3)+(0,0.2)$);
\coordinate (s4) at (-1,0);
\coordinate (t4) at ($(s4)+(0,0.2)$);
\coordinate (s5) at (1.3,0);
\coordinate (t5) at ($(s5)+(0,0.2)$);
\coordinate (s6) at (3,0);
\coordinate (t6) at ($(s6)+(0,0.2)$);
\coordinate (s7) at (4.5,0);
\coordinate (t7) at ($(s7)+(0,0.2)$);
\coordinate (s8) at (6,0);
\coordinate (t8) at ($(s8)+(0,0.2)$);

% draw ticks
\draw [line width=1.5pt, gray] (s0) -- (t0);
\node [anchor=south] at (-10, 0.6) {\texttt{ARIMA}};
\node [anchor=south] at (t0.north) {\texttt{LSTM DT}};
\node [anchor=north] at (s0.south) {\texttt{0.1.3}};

\draw [line width=1.5pt, gray] (t1) -- (s1);
\node [anchor=south] at (t1.north) {\texttt{Azure AD}};
\node [anchor=north] at (s1.south) {\texttt{0.1.4}};

\draw [line width=1.5pt, gray] (t2) -- (s2);
\node [anchor=south] at (t2.north) {\texttt{TadGAN}};
\node [anchor=north] at (s2.south) {\texttt{0.1.5}};

\draw [line width=1.5pt, gray] (t3) -- (s3);
\node [anchor=south] at (-4, 0.6) {\texttt{LSTM AE}};
\node [anchor=south] at (t3.north) {\texttt{Dense AE}};
\node [anchor=north] at (s3.south) {\texttt{0.1.6}};

\draw [line width=1.5pt, gray] (t4) -- (s4);
\node [anchor=south] at (t4.north) {\texttt{AER}};
\node [anchor=north] at (s4.south) {\texttt{0.3.1}};

\draw [line width=1.5pt, gray] (t5) -- (s5);
\node [anchor=south] at (t5.north) {\texttt{VAE}};
\node [anchor=north] at (s5.south) {\texttt{0.4.0}};

\draw [line width=1.5pt, gray] (t6) -- (s6);
\node [anchor=south] at (t6.north) {\texttt{GANF}};
\node [anchor=north] at (s6.south) {\texttt{0.4.1}};

\draw [line width=1.5pt, gray] (t7) -- (s7);
\node [anchor=south] at (t7.north) {\texttt{MP}};
\node [anchor=north] at (s7.south) {\texttt{0.5.2}};

\draw [line width=1.5pt, gray] (t8) -- (s8);
\node [anchor=south] at (t8.north) {\texttt{LNN}};
\node [anchor=north] at (s8.south) {\texttt{0.6.0}};

\node [anchor=north] at (-3, -0.6) {Version Release};
\end{tikzpicture}
    \vspace{-0.8cm}
    \caption{Timeline of pipeline introduction to the benchmark. \bench started with 2 pipelines; over the course of three years, 8 more pipelines were introduced at different stages.}
    % \vspace{-0.5cm}
    \label{fig:timeline}
\end{figure*}

% These shifts are highlighted in Table~\ref{tab:change} that shows the percentage change in scores.
% We define a shift when the percentage of change is $\mu \geq 1$ and/or $\delta \geq 2$ where $\mu$ is the average percentage change in F1 score performance between consecutive releases, and $\delta$ is the standard deviation.
First, in version update \texttt{0.1.3 $\rightarrow$ 0.1.4}, we saw a drop in F1 score due to an internal change in how we calculate the overall scores.
The aggregation calculation became automated and was conducted on the dataset level rather than the signal level.
Second, in version update \texttt{0.1.5 $\rightarrow$ 0.1.6}, there was an increase in performance that can be traced back to our hyperparameter setting modifications.
% , including making them dependent on signal length.
Third, going from version \texttt{0.3.2 $\rightarrow$ 0.4.0} shifted our implementation from \texttt{TensorFlow} version 1 to 2, which impacted the underlying implementation.
Lastly, after introducing a new dataset, namely UCR, we noticed a drop in the overall performance by pipelines in \texttt{0.5.0 $\rightarrow$ 0.5.1} because this was a more difficult dataset.
% For Orion developers, running the benchmark every release reassures that no performance disruption has happened.
% Moreover, it stabilizes pipelines and makes sure that they do not go out of date, especially with dependency package updates.
Overall, the observed changes were minimal and could be traced back to alterations within our framework.

\textbf{Pipeline Integration.}
Figure~\ref{fig:timeline} showcases exactly when each pipeline was integrated to \bench.
The first benchmark release, version \texttt{0.1.3}, in September 2020, featured only 2 pipelines.
Over time, new models have been developed and integrated.
As of today, \bench has 12 verified pipelines, ranging from classical models to deep learning and made by 5 different contributors.

\definecolor{mygray}{gray}{0.8}
\newcommand{\cmark}{\ding{51}}
\newcommand{\xmark}{\color{mygray}\ding{55}}     
    
\begin{table*}[!t]
\centering
\caption{Comparison of anomaly detection benchmarks. A (\ding{51}) indicates the framework includes an attribute, while an ({\color{mygray}\ding{55}}) indicates the attribute is absent. %
\underline{\smash{\#Datasets}} and \underline{\smash{\#Pipelines}} columns represent the number of currently available datasets and pipelines respectively. % 
Columns under ``Pipeline Type" represent whether certain pipeline types are supported including \underline{\smash{classic}} pipelines such as ARIMA, Deep Learning (\underline{\smash{DL}}) pipelines such as LSTM, and BlackBox (\underline{\smash{BBox}}) pipelines that are called externally through an API such as Azure's AD service. %
Columns under ``Properties" represent whether a benchmark has certain properties, including custom \underline{\smash{evaluation}} methods for time series anomaly detection; %
Whether the benchmark is \underline{\smash{extensible}} and can integrate new datasets and pipelines, and %
If the benchmark is being released in \underline{\smash{periodic}} fashion with an updated leaderboard. %
The last two columns illustrate the last time a leaderboard has been published and where.}
% whether it has been shared in a paper or directly on github.}
\resizebox{\textwidth}{!}{\begin{tabular}{l*{14}{c}}
\toprule
% header
        & \multicolumn{2}{c}{Available} & \multicolumn{3}{c}{Pipeline Type} & \multicolumn{3}{c}{Properties} & \multicolumn{2}{c}{Published Leaderboard} \\
        \cmidrule(lr){2-3}\cmidrule(lr){4-6}\cmidrule(lr){7-9}\cmidrule(lr){10-11}
\textbf{Framework}  & \# Datasets   &   \# Pipelines    &   Classic   & DL   & BBox   &   Evaluation  & Extensible    & Periodic  & Last Update    & Source  \\
\midrule
\texttt{Numenta}\hfill~\cite{lavin2015numentabenchmark}    & 7                   & 4     & \cmark & \xmark & \xmark & \cmark    & \cmark    & \xmark &   Jun 2018        & Github \\
\texttt{TSB-UAD}\quad~\cite{paparrizos2022tsb-uad}         & 18                  & 12    & \cmark & \cmark & \xmark & \cmark    & \xmark    & \xmark &   Nov 2022        & Github \\
\texttt{TODS}\hfill~\cite{lai2021todsbenchmark}            & 4                   & 9     & \cmark & \cmark & \xmark & \xmark    & \xmark    & \xmark &   Dec 2021        & Paper \\
\texttt{TimeEval}\hfill~\cite{wenig2022timeeval}           & 23                  & 71    & \cmark & \cmark & \xmark & \xmark    & \cmark    & \xmark &   Aug 2022        & Paper \\
\texttt{Exathlon}\hfill~\cite{jacob2020exathlon}           & 10                  & 3     & \xmark & \cmark & \xmark & \cmark    & \xmark    & \xmark &   Sep 2021        & Paper \\
\texttt{Merlion}\hfill~\cite{bhatnagar2021merlion}         & 12                  & 12    & \cmark & \cmark & \xmark & \cmark    & \xmark    & \xmark &   Sep 2021        & Paper \\
\midrule
\textbf{\bench}                                            & 14                  & 12    & \cmark & \cmark & \cmark & \cmark    & \cmark    & \cmark &   Oct 2024        & Github \\
\bottomrule
\end{tabular}}
\label{tab:related_work}
\vspace{-1em}
\end{table*}

\subsection{OrionBench in Action}
\label{sec:action}
As anomaly detection models continue to be developed, \bench allows researchers and end-users to understand and compare these models.
In this subsection, we walk through two real-world scenarios where benchmarking was useful for:
(1) guiding researchers to develop a new model for unsupervised time series anomaly detection;
(2) providing end-users with an existing SOTA model.
We show that \bench is a commodity benchmarking framework.

Before walking through the aforementioned scenarios, we would like to describe the state of OrionBench. \texttt{LSTM DT}~\cite{hundman2018lstmdt} and \texttt{TadGAN}~\cite{geiger2020tadgan} (which was developed by the Orion team) performed competitively against each other until version \texttt{0.3.1}, when \texttt{AER}~\cite{wong2022aer} was introduced.
Below, we illustrate the story behind the \texttt{AER} model and how we, the OrionBench developers, helped benchmark their model.

\noindent\textbf{Scenario 1 -- OrionBench guided a researcher to focus in the right direction.}
% Researchers are inspired by the latest innovations in deep learning.
% They are eager to adopt them for new tasks and explore their capabilities.
Researchers are eager to adopt the latest innovations in deep learning. 
An independent researcher was keen on introducing the attention mechanism to anomaly detection~\cite{vaswani2017attention}.
While the model was promising in local experiments, to assure its performance we decided to run it through \textit{OrionBench}. Unfortunately, the model could not do better than either \texttt{LSTM DT} or \texttt{TadGAN}.
This reoriented the project and led to an investigation of the successes and limitations of pipelines.
Subsequently, it led to a deep understanding of where prediction models prevailed compared to reconstruction models and vice versa.
% Part of our endeavour is to make model development an easier process.
% Suppose that you are interested in developing your own model, where would you start from?
% The story behind~\citet{wong2022aer} involved a deep understanding of where prediction models prevailed compared to reconstruction models and vice versa.
\bench helped guide this process by cross-referencing model performance with dataset properties.
The conclusion was that prediction-based anomaly scores are better at capturing point anomalies than reconstruction-based anomaly scores.
Moreover, reconstruction-based anomaly scores are better at capturing longer anomalies.
\citet{wong2022aer} uncovered more associations related to anomaly scores and error methods. 
The outcome of this investigation ultimately resulted in the \texttt{AER} model and is now the best-performing pipeline on \textit{OrionBench}.\\
% a better-performing model, which was published as

\noindent\textbf{Scenario 2 -- OrionBench enabled the addition of a latest model and provided an end-user with confidence in other models.}
We had been working with an end-user from a renowned satellite company for over four years when they approached us with interest in a new SOTA model.
The model was \texttt{GANF},~\cite{dai2022graph} which had been featured in a news article~\footnote{https://news.mit.edu/2022/artificial-intelligence-anomalies-data-0225} that caught their attention.
New models are published that claim SOTA performance, beating existing models on their benchmarks.
This is common for renowned companies that invest in creating high-performing models.
The end-user wanted to know: \textit{Should we adopt this model?}
Several issues can prevent such models from living up to their promised performance in industrial and operational settings.
Real-world datasets are inherently more complex than pristine benchmark datasets.
Furthermore, authors often fine-tune a model to the benchmark datasets, neglecting others and causing their model to underperform on unseen datasets.
\textit{OrionBench}, as an independent benchmark, can help determine whether it makes sense to adopt a new model.
We integrated \texttt{GANF} into \textit{OrionBench}. As presented earlier in Figure~\ref{fig:f1score}, it was only competitive on the NAB dataset.
However, due to the seamless integration of the pipelines into \bench, the end-user was still able to apply the pipeline to their own data and obtained valuable results.
This emphasizes that the behavior of models differs from one dataset to another, and there is no one-pipeline-fits-all.
% We include pipelines from many companies including: NASA~\cite{hundman2018lstmdt}, IBM~\cite{dai2022graph}, and Microsoft~\cite{ren2019azure}.
% Although we currently only publish the leaderboard on publicly available datasets, users have the option of running the benchmark on their own custom data.

Similarly, \texttt{LNN} models~\cite{hasani2021lnn} have been utilized in a variety of applications, including robot control. A published news article\footnote{https://news.mit.edu/2021/machine-learning-adapts-0128} suggests that these models are able to perform any time series task.
To test their ability to perform unsupervised anomaly detection, we implemented an \texttt{LTC} primitive and, shortly after, the \texttt{LNN} pipeline.
\citet{hasani2021lnn} released an accompanying \texttt{pip} installable library, which has made creating the \texttt{LNN} pipeline straightforward. 
It took one week from its first commit to when it merged on the main branch and became sandbox-available.
% \texttt{LNN} performs well in the benchmark; however, its main drawback is its computational time.
\bench has made it easier for us to incorporate new models and assess their anomaly detection capabilities.

% \textcolor{blue}{How do we deal with LNN and GANF if they end up deleting their code or making it private all of the sudden?
% Managing PyPI and what are the challenges in that?}

% \footnotetext[1]{Additional synthetic and artificial data is generated in the benchmark.}
% \footnotetext[2]{Data are traces from 10 distributed streaming jobs on a Spark cluster.}

\vspace{-0.3em}
\section{Related Work}
\vspace{-0.3em}
\label{sec:related}

In this section, % we focus on time series anomaly detection benchmarks. 
we walk-through some of the algorithms on unsupervised time series anomaly detection as well as benchmarking systems. % in Appendix~\ref{app:related}.
\vspace{-0.5em}

\subsection{Unsupervised Time Series Anomaly Detection Algorithms}
Many anomaly detection methods have emerged in the past few years~\cite{chandola2009survey, blazquez2021survey2, goldstein2016survey1}. These include
statistical thresholding techniques~\cite{patcha2007statisticalthreshold}, clustering-based methods~\cite{munz2007clustering1, syarif2012clustering2, agrawal2015clustering3}, and machine learning models~\cite{hasan2019attack(ml), liu2015opprentice(ml)}.
More recently, deep learning models have become popular and have been adopted for anomaly detection~\cite{chalapathy2019survey3, pang2021survey4}.
Deep learning-based anomaly detection models for time series data rely mostly on unsupervised learning, because in most settings,  there is no a priori knowledge of anomalous events.
\citet{malhotra2016lstmae} built an autoencoder with Long Short-Term Memory (LSTM) layers~\cite{hochreiter1997longshorttermmemory} that learns to reconstruct `normal' signal behavior. 
It uses the residual between the reconstructed signal and the original signal to locate anomalies.
LSTM networks are practical at capturing the temporal dynamics in time series data.
\citet{hundman2018lstmdt} used an LSTM forecasting model to predict the signal and paired it with a non-parametric threshold to mitigate false positives.
Since then, generative models including variational autoencoders (VAE)~\cite{park2018vae}, Generative Adversarial Networks (GAN)~\cite{geiger2020tadgan}, and Transformers~\cite{xu2022anomalytransformer} have been adopted for unsupervised anomaly detection.

\subsection{Time Series Anomaly Detection Benchmarks}
There are several notable open-source time series benchmarking systems featuring unsupervised time series anomaly detection methods~\cite{lavin2015numentabenchmark, paparrizos2022tsb-uad, lai2021todsbenchmark, wenig2022timeeval, jacob2020exathlon, bhatnagar2021merlion}.
Table~\ref{tab:related_work} highlights the key features present in each framework.
% time series anomaly detection benchmarks have been introduced~\cite{jacob2020exathlon, lai2021todsbenchmark, paparrizos2022tsb-uad, lavin2015numentabenchmark, wenig2022timeeval}, 
While these benchmarks do exist, they are not directed towards \user{s}, and are usually not kept up-to-date. Table~\ref{tab:related_work} shows the latest published results for each time series anomaly detection benchmark framework, whether the scoreboard has been updated on github, and the date of the latest update. Usually, these assessments are done once, during the production of related papers. We argue that this makes them \textit{point-in-time} benchmarking frameworks. 
% Further comparisons are addressed in Section~\ref{sec:related}.

% There are several notable open-source time series benchmarking systems featuring unsupervised time series anomaly detection methods~\cite{lavin2015numentabenchmark, paparrizos2022tsb-uad, lai2021todsbenchmark, wenig2022timeeval, jacob2020exathlon, bhatnagar2021merlion}.
% Table~\ref{tab:related_work} highlights the key features present in each framework.
In addition to unsupervised pipelines, some frameworks include supervised pipelines~\cite{paparrizos2022tsb-uad, wenig2022timeeval}. 
However, comparing supervised pipelines to unsupervised ones can be misleading, as labels are not available in most real-world scenarios.
We address three key points with \textit{OrionBench}:
(1) time series anomaly detection requires careful consideration of how to evaluate pipelines;
(2) integration of new pipelines and datasets needs to be seamless such that pipelines are available to \user{s},
(3) benchmarks need periodic releases and leaderboard updates to ensure results are trusted and pipelines are stable.

There are many other benchmarking frameworks, such as time series forecasting benchmarks~\cite{gluonts_jmlr, bauer2021libra, taieb2012reviewforecasting}, and anomaly detection for tabular data~\cite{campos2016tabularevaluation, han2022adbench}.
However, these benchmarks inherently differ from our unsupervised anomaly detection benchmark for time series data.

\section{Conclusion}
\label{sec:discussion}

We present \bench -- a continuous end-to-end benchmarking framework for unsupervised time series anomaly detection. 
The benchmark is open-source and publicly available: \url{https://github.com/sintel-dev/Orion}.
As of today, the benchmark holds 28 primitives, 12 pipelines, 14 public datasets, and 2 custom evaluation metrics.
We present the qualitative and computational performance of pipelines across all datasets.
Moreover, we showcase results our benchmark has accumulated since 2020, highlighting its value for providing continuous evaluations that demonstrate the extensibility and stability of pipelines.

Although the benchmark compares different pipelines, there is no one pipeline will be the best choice for every dataset.
Pipeline selection is still a challenging process that highly correlates with the characteristics of the dataset at hand and the type of anomalies present in the dataset.
In our future work, we would like to focus on the suitability of pipelines and finding the relationship between various attributes of the input data and the efficacy of anomaly detection.
Moreover, we invite ML researchers to contribute to \textit{OrionBench}.

\section*{Acknowledgements}
This research was supported by SES S.A., and Iberdrola and ScottishPower Renewables.

\section*{Change Log}
Updated the title; 
Table~\ref{tab:pprq} has been updated and now gives more details.
Figure~\ref{fig:elapsed} now shows runtime groupby by three different signal length bins;
Added release 0.6.1 to the results.

\bibliography{ref}
\bibliographystyle{icml2023}

%%%%%%%%%%%%%%%%%%%%%%%%%%%%%%%%%%%%%%%%%%%%%%%%%%%%%%%%%%%%%%%%%%%%%%%%%%%%%%%
%%%%%%%%%%%%%%%%%%%%%%%%%%%%%%%%%%%%%%%%%%%%%%%%%%%%%%%%%%%%%%%%%%%%%%%%%%%%%%%
% APPENDIX
%%%%%%%%%%%%%%%%%%%%%%%%%%%%%%%%%%%%%%%%%%%%%%%%%%%%%%%%%%%%%%%%%%%%%%%%%%%%%%%
%%%%%%%%%%%%%%%%%%%%%%%%%%%%%%%%%%%%%%%%%%%%%%%%%%%%%%%%%%%%%%%%%%%%%%%%%%%%%%%
    
\newpage
\appendix
\onecolumn
\section*{Appendix}
\section{Limitations}
We acknowledge several limitations of our framework and results.
Black box pipelines such as Microsoft's \texttt{Azure AD} service lack certain levels of transparency.
In Figure~\ref{fig:version}, we noticed that \texttt{Azure AD} improved in \texttt{0.1.6 $\rightarrow$ 0.1.7}.
However, we have no knowledge on what has caused this improvement.
Unlike other pipelines where we can cross reference change of behaviour to code modification or even updated dependency package releases.
Nevertheless, we are still able to monitor the performance of black box pipelines and having some confidence in their stability.

In addition, benchmarks are notorious for requiring massive computing resources, and in this case it is no different.
While the models can vary in usage, to perform a comprehensive benchmark, we utilize MIT supercloud~\cite{reuther2018supercloud}.
When computing resources are limited, on-demand benchmark runs become difficult.
We aim to alleviate this challenge with continuous and periodic running benchmarks.
Moreover, with every introduction to a new model, a benchmark must be run to add the model to the leaderboard.

Lastly, and most importantly, there is no guarantee that these pipelines will deliver the same performance on real-world datasets. A clear example was demonstrated in Section~\ref{sec:action} Scenario 2 where \texttt{GANF} produced valuable results for the end-user, however its results in the benchmark are not as promising as some other pipelines. This stresses the importance of pipeline selection based on the characteristics of the datasets and anomalies. Further research is needed to understand the suitability of unsupervised pipelines for a given dataset.

\section{Primitives \& Pipelines}
\subsection{Primitive Template}

Abstractions in \bench of primitives and pipelines are universal representations of end-to-end models, from a signal to a set of detected anomalies. 
Compared to standard \texttt{scikit-learn} like code, it requires one additional step of creating \texttt{json} files to define these primitives.
Figure~\ref{fig:primitive_template} showcases a template that helps contributors to guides their own primitive.

Once primitives are built, they can be stacked to create a pipeline similar to the example shown in Figure~\ref{fig:lstm_pipeline_example}.
The example shows the \texttt{json} file representation of \texttt{LSTM DT} pipeline.

\begin{figure}
    \centering
\begin{minted}{python}
{
    "name": "orion.primitives.primitive.PrimitiveName",
    "contributors": ["Author <email>"],
    "documentation": "reference to documentation or paper if available.",
    "description": "short description.",
    "classifiers": {
        "type": "postprocessor",
        "subtype": "anomaly_detector"
    },
    "modalities": [],
    "primitive": "orion.primitives.primitive.PrimitiveName",
    "fit": {
        "method": "fit",
        "args": [
            {
                "name": "X",
                "type": "ndarray"
            },
            {
                "name": "y",
                "type": "ndarray"
            }
        ]
    },
    "produce": {
        "method": "detect",
        "args": [
            {
                "name": "X",
                "type": "ndarray"
            },
            {
                "name": "y",
                "type": "ndarray"
            }
        ],
        "output": [
            {
                "name": "y",
                "type": "ndarray"
            }
        ]
    },
    "hyperparameters": {
        "fixed": {
            "hyper_name": {
                "type": "str",
                "default": "value"
            }
        }
    }
}
\end{minted}
    \caption{Primitive template. The first section of the \texttt{json} describes metadata, the second part contains functional information including the names of the methods and their arguments, the third part defines the hyperparameters of the primitive.}
    \label{fig:primitive_template}
\end{figure}

\begin{figure}
    \centering
\begin{minted}{python}
{
    "primitives": [
        "mlstars.custom.timeseries_preprocessing.time_segments_aggregate",
        "sklearn.impute.SimpleImputer",
        "sklearn.preprocessing.MinMaxScaler",
        "mlstars.custom.timeseries_preprocessing.rolling_window_sequences",
        "keras.Sequential.LSTMTimeSeriesRegressor",
        "orion.primitives.timeseries_errors.regression_errors",
        "orion.primitives.timeseries_anomalies.find_anomalies"
    ],
    "init_params": {
        "mlstars.custom.timeseries_preprocessing.time_segments_aggregate#1": {
            "time_column": "timestamp",
            "interval": 21600,
            "method": "mean"
        },
        "sklearn.preprocessing.MinMaxScaler#1": {
            "feature_range": [
                -1,
                1
            ]
        },
        "mlstars.custom.timeseries_preprocessing.rolling_window_sequences#1": {
            "target_column": 0,
            "window_size": 250
        },
        "keras.Sequential.LSTMTimeSeriesRegressor": {
            "epochs": 35
        },
        "orion.primitives.timeseries_anomalies.find_anomalies#1": {
            "window_size_perc": 30,
            "fixed_threshold": false
        }
    }
}
\end{minted}
    \caption{\texttt{LSTM DT} pipeline example. This is the content present in the \texttt{json} file of the pipeline. The first section defines the stack of primitives used in the pipeline which will be computed to the graph shown previously in Figure~\ref{fig:pipeline}. The \texttt{init} argument initializes some of the hyperparameters for each primitive. This is a detailed version with full primitive names of the hyperparameters shown in Figure~\ref{fig:hyper_config}.}
    \label{fig:lstm_pipeline_example}
\end{figure}

\subsection{Pipelines}
Currently in \textit{OrionBench}, there are 9 readily available pipelines.
They are all unsupervised pipelines.
All pipelines and their hyperparameter settings for the benchmark can be explored directly: \url{https://github.com/sintel-dev/Orion/tree/master/orion/pipelines/verified}.
Below we provide further detail on the mechanisms behind each pipeline.

\noindent\texttt{ARIMA}~\cite{pena2013arima}.
\texttt{ARIMA} is an autoregressive integrated moving average model which is a classic statistical analysis model.
It is a forecasting model that learns autocorrelations in the time series to predict future values prediction. 
Since then it has been adapted for anomaly detection.
The pipeline computes the prediction error between the original signal and the forecasting one using simple point-wise error.
Then it pinpoints where the anomalies are based one when the error exceeds a certain threshold.
Particularly, \texttt{ARIMA} pipeline uses a moving window based thresholding technique defined in \texttt{find\_anomalies} primitive.

\noindent\texttt{AER}~\cite{wong2022aer}.
\texttt{AER} is an autoencoder with regression pipeline.
It combines prediction and reconstruction models simultaneously. 
More specifically, it produces bi-directional predictions (forward \& backward) while reconstructing the original time series at the same time by optimizing a joint objective function.
The error is then computed as a point-wise error for both forward and backward predictions.
As for reconstruction, dynamic time warping is used, which computes the euclidean distance between two time series where one might lag behind another.
The total error is then computed as a point-wise product between the three aforementioned errors.

\noindent\texttt{LSTM DT}~\cite{hundman2018lstmdt}.
\texttt{LSTM DT} is a prediction-based pipeline using an LSTM model.
Similar to \texttt{ARIMA}, it computes the residual between the original signal and predicted one using smoothed point-wise error.
Then they apply a non-parametric thresholding method to reduce the amount of false positives.

\noindent\texttt{TadGAN}~\cite{geiger2020tadgan}.
\texttt{TadGAN} is a reconstruction pipeline that uses generative adversarial networks to generate a synthetic time series.
To sample a ``similar'' time series, the model uses an encoder to map the original time series to the latent dimension.
There are three possible strategies to compute the errors between the real and synthetic time series. 
Specifically, point-wise errors, area difference, and dynamic time warping.
Most datasets are set to dynamic time warping (dtw) as error.

\noindent\texttt{MP}~\cite{yeh2016matrixprofile}.
\texttt{MP} is a matrix profile method that seeks to find discords in time series.
The pipeline computes the matrix profile of a signal, which essential provides the closes nearest neighbor for each segment.
Based on these values, segments with large distance values to their nearest neighbors are anomalous.
We use \texttt{find\_anomalies} to set the threshold dynamically.

\noindent\texttt{VAE}~\cite{park2018vae}.
\texttt{VAE} is a variational autoencoder consisting of an encoder and a decoder with LSTM layers.
Similar to previous pipelines, it adopts reconstruction errors to compute the deviation between the original and reconstructed signal.

\noindent\texttt{LSTM AE}~\cite{malhotra2016lstmae}.
\texttt{LSTM AE} is an autoencoder with an LSTM encoder and decoder.
This is a simpler variant of \texttt{VAE}.
It also uses reconstruction errors to measure the difference between the original and reconstruction signal.

\noindent\texttt{Dense AE}.
\texttt{Dense AE} is an autoencoder where its properties are exactly similar to that of \texttt{LSTM AE} with the exception of the encoder and decoder layers.

\noindent\texttt{GANF}~\cite{dai2022graph}.
\texttt{GANF} is density-based methods where they use normalizing flows to learn the distribution of the data with a graph structure to overcome the challenge of high dimensionality.
The model outputs an \textit{anomaly measure} that indicates where the anomalies might be.
To convert the output into a list of intervals, we add \texttt{find\_anomalies} primitive.

\noindent\texttt{Azure AD}~\cite{ren2019azure}.
\texttt{Azure AD} is a black box pipeline which connects to Microsoft's anomaly detection service~\footnote{\url{https://azure.microsoft.com/en-us/products/cognitive-services/anomaly-detector/}}.
To use this pipeline, the user needs to have a subscription to the service.
Then the user can update the \texttt{subscription\_key} and \texttt{endpoint} in the pipeline \texttt{json} for usage.

\noindent\texttt{AnomTransformer (AT)}~\cite{xu2022anomalytransformer}.
\texttt{AnomTransformer} is a transformer based model using a new \textit{anomaly-attention} mechanism to compute the association discrepancy.
The model amplifies the discrepancies between normal and abnormal time points using a minimax strategy.
The threshold is set based on the attention values.

\section{Data}

\subsection{Data Format}
Time series is a collection of data points that are indexed by time. There are many forms in which time series can be stored, we define a time series as a set of time points, which we represent through integers denoting \textit{timestamps}, and a corresponding set of values observed at each respective timestamp.
Note that no prior pre-processing is required as all pre-processing steps are part of the pipeline, e.g. imputations, scaling, etc.

\subsection{Dataset Details}
The benchmark currently features 11 publicly accessible datasets from different sources.
Table~\ref{tab:datasets} illustrates some of the datasets' properties.
Below, we provide more detailed description for each dataset.

\noindent\texttt{NASA}.
This dataset is a spacecraft telemetry signals provided by NASA.
It was originally released in 2018 as part of the \textsc{Lstm-DT} paper~\cite{hundman2018lstmdt} and can be accessed directly from \url{https://github.com/khundman/telemanom}.
It features two datasets: Mars Science Laboratory (MSL) and Soil Moisture Active Passive (SMAP).
MSL contains 27 signals with 36 anomalies.
SMAP contains 53 signals with 69 anomalies.
In total, NASA datasets has 80 signals with 105 anomalies.
This dataset was pre-split into training and testing partitions.
In our benchmark, we train the pipeline using the training data, and apply detection to only the testing data.

\noindent\texttt{NAB}.
Part of the Numenta benchmark~\cite{lavin2015numentabenchmark} is the NAB dataset \url{https://github.com/numenta/NAB}.
This datasets includes multiple types of time series
data from various applications and domains 
In our benchmark we selected five sub-datasets (name: \# signals, \# anomalies):
artWithAnomaly (\texttt{Art}: 6, 6): this dataset was artificially generated;
realAWSCloudwatch (\texttt{AWS}: 17, 20): this dataset contains AWS server metrics collected by AmazonCloudwatch service such as CPU Utilization;
realAdExchange (\texttt{AdEx}: 5, 11),  this dataset contains online advertisement clicking rate metrics such as cost-per-click;
realTraffic (\texttt{Traf}: 7, 14): this dataset contains real time traffic metrics from the Twin Cities Metro area in Minnesota such as occupancy, speed, etc;
and realTweets (\texttt{Tweets}: 10, 33): this dataset contains metrics of a collection of Twitter mentions of companies (e.g. Google) such as number of mentions each 5 minutes.

\noindent\texttt{Yahoo S5}.
This dataset contains four different sub-datasets.
A1 dataset is based on real production traffic of Yahoo computing systems with 67 signals and 179 anomalies.
On the other hand, A2, A3 and A4 are all synthetic datasets with 100 signals each and 200, 939, and 835 anomalies respectively.
There are many anomalies in this dataset with over 2,153 in 367 signals, averaging 5.8 anomalies in each signal.
Most of the anomalies in A3 and A4 are short and last for only a few points in time.
Data can be requested from Yahoo's website \url{https://webscope.sandbox.yahoo.com/catalog.php?datatype=s&did=70}.
In our benchmark, we train and apply detection to the same entire signal.

\noindent\texttt{UCR}.
This dataset was released in a SIGKDD competition in 2021 \url{https://www.cs.ucr.edu/~eamonn/time_series_data_2018/UCR_TimeSeriesAnomalyDatasets2021.zip}.
It contains 250 signals with only one anomaly in each signal.
The anomalies themselves were artificially introduced to the signal.
More specifically, in many times they are synthetic anomalies, or a consequence of flipping/smoothing/interpolating/reversing/prolonging normal segments to create anomalies.
The dataset was created to have more challenging cases of anomalies.

\section{Evaluation}
This section provides further details on our evaluation setup and obtained results.
Code to reproduce Figures and Tables are provided \url{https://github.com/sarahmish/orionbench-paper}

\subsection{Evaluation Setup}
Results presented in Section~\ref{sec:demo} are reported based on version \texttt{0.5.0} of \href{https://github.com/sintel-dev/Orion}{Orion} which is also released on pip~\footnote{\url{https://pypi.org/project/orion-ml/0.5.0/}}.
We recommend setting up a new python environment before installing Orion.
Currently the library is supported in \texttt{python 3.6}, \texttt{3.7}, and \texttt{3.8}.

\noindent\textbf{Evaluation Strategy.}
Measuring the performance of unsupervised time series anomaly detection pipelines is more nuanced than the usual classification metrics.
\bench compares detected anomalies with ground truth labels according to well-defined metrics.
This can be done using either weighted segment or overlapping segment~\cite{alnegheimish2022sintel}.
For our evaluation in this paper, we use \textit{overlapping segment} exclusively.
Using overlapping segment, for each experiment run (which is an evaluation of one pipeline on one signal), we record the number of true positives (TP), false positive (FP), and false negative (FN) obtained.
Because anomalies are scarce and in many signals only one anomaly exists (or none), in many cases precision and recall scores will be undefined on a signal level.
Therefore, we compute the scores on a dataset level.

\[
precision = \frac{\sum_{s \in \mathcal{S}} TP_s}{\sum_{s \in \mathcal{S}} TP_s + FP_s}
\qquad
recall = \frac{\sum_{s \in \mathcal{S}} TP_s}{\sum_{s \in \mathcal{S}} TP_s + FN_s}
% \qquad
% f1 = \frac{\sum_{s \in \mathcal{S}} TP_s}{\frac{1}{2} \sum_{s \in \mathcal{S}} FP_s + FN_s}
\]

For a given dataset with a set of signals $\mathcal{S}$, we compute the total true positives, false positives, and false negatives within every signal in that set.
Then we compute the score for each pipeline according to the metric of interest whether it is precision, recall, or f1 score. The computation of f1 score is standard from precision and recall ($f1 = 2 \times \frac{precision \times recall}{precision + recall}$).

\noindent\textbf{Recorded Information.}
During the benchmark process, information regarding performance, computation, diagnostics, etc. gets recorded.
Below we list all the information we store for each experiment.
An experiment is defined as a single pipeline trained on a single time series then used for detection for the same time series.

\begin{itemize}
    \item \textit{dataset:} the dataset that the signal belongs to, e.g. \texttt{SMAP}.
    \item \textit{pipeline:} the name of the pipeline, e.g. \texttt{AER}.
    \item \textit{signal:} the name of the signal, e.g. \texttt{S-1}.
    \item \textit{iteration:} each experiment can be run for $k$ iterations.
    \item \textit{f1, precision, recall:} the evaluated metrics, in many cases it is undefined.
    \item \textit{tn, fp, fn, tp:} the evaluated number of true negatives, false positives, false negatives, true positives respectively. In overlapping segment approach, \textit{tn} does not have a value given the nature of evaluation.
    \item \textit{status:} whether or not the experiment ran from beginning to end without issue.
    \item \textit{elapsed:} how much runtime each experiment took (includes training and inference).
    \item \textit{run\_id:} the process identification number.
\end{itemize}

The benchmark results are saved as \texttt{.csv} files and stored directly in the Github repositories: \url{https://github.com/sintel-dev/Orion/tree/master/benchmark/results}.
Moreover, the pipelines used in each experiment are saved for reproduciblity measures.
Due to their large size, we store these pipelines on a local server.
However, part of our endeavour is to make these pipelines public as well such that they can be used and inspected.

\begin{figure}[ht]
    \centering
    \includegraphics[width=1\linewidth]{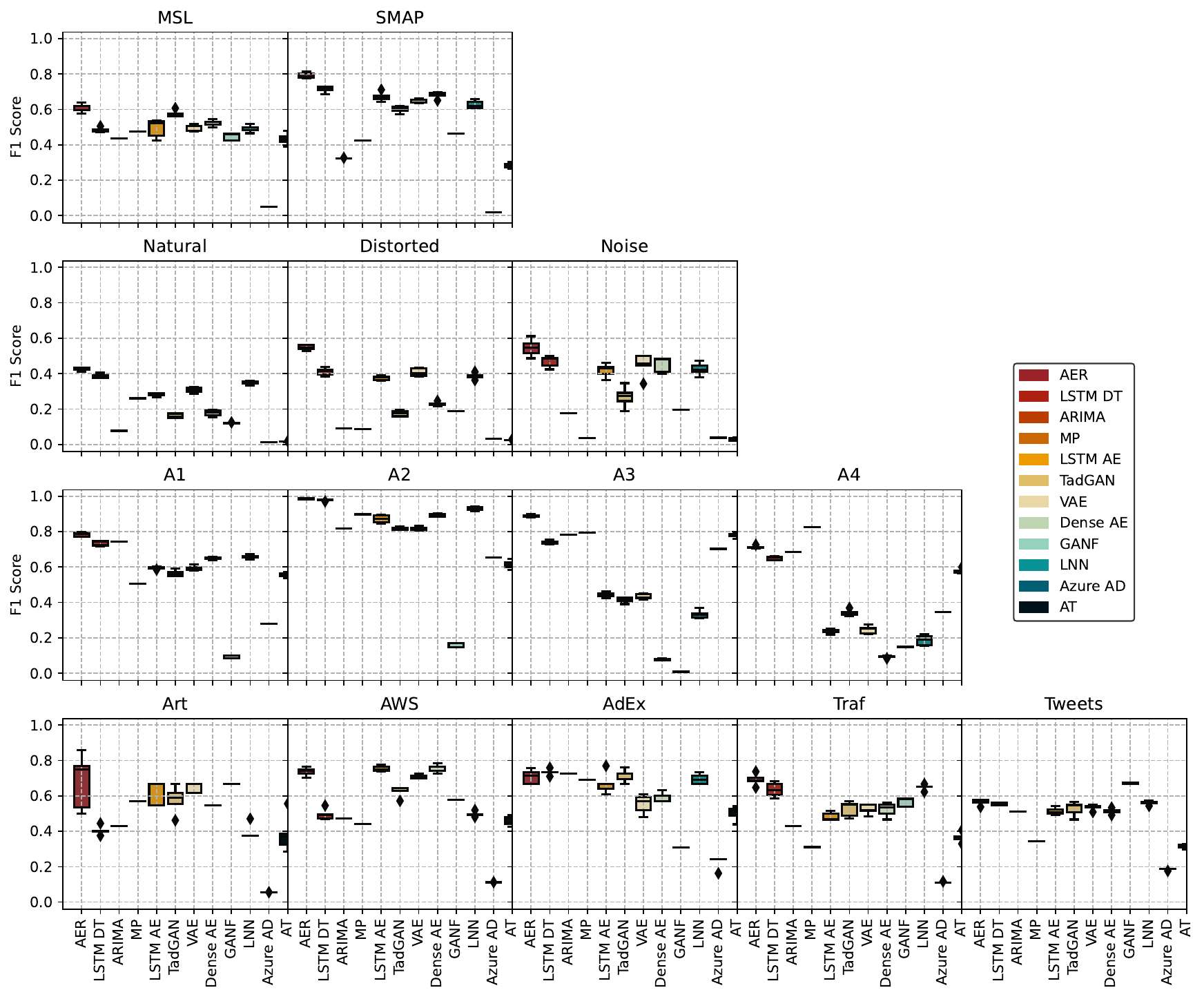}
    \caption{Distribution of F1 Score per Dataset. This Figure is a detailed version of Figure~\ref{fig:f1score} where now every dataset is shown separately. On average, \texttt{AER} is the highest scoring pipeline for most datasets with the exception of AWS, AdEx, and Tweets. The performance of pipelines changes from one dataset to another, indicating there does not exists a single pipeline that will work perfectly for all datasets. One of the insights we find is how point anomalies in A3 \& A4 present a challenge for reconstruction-based pipelines such as \texttt{LSTM AE}, \texttt{TadGAN}, \texttt{VAE}, and \texttt{Dense AE}.}
    \label{fig:f1-boxplot-dataset}
\end{figure}

\begin{figure}[ht]
    \centering
    \includegraphics[width=1\linewidth]{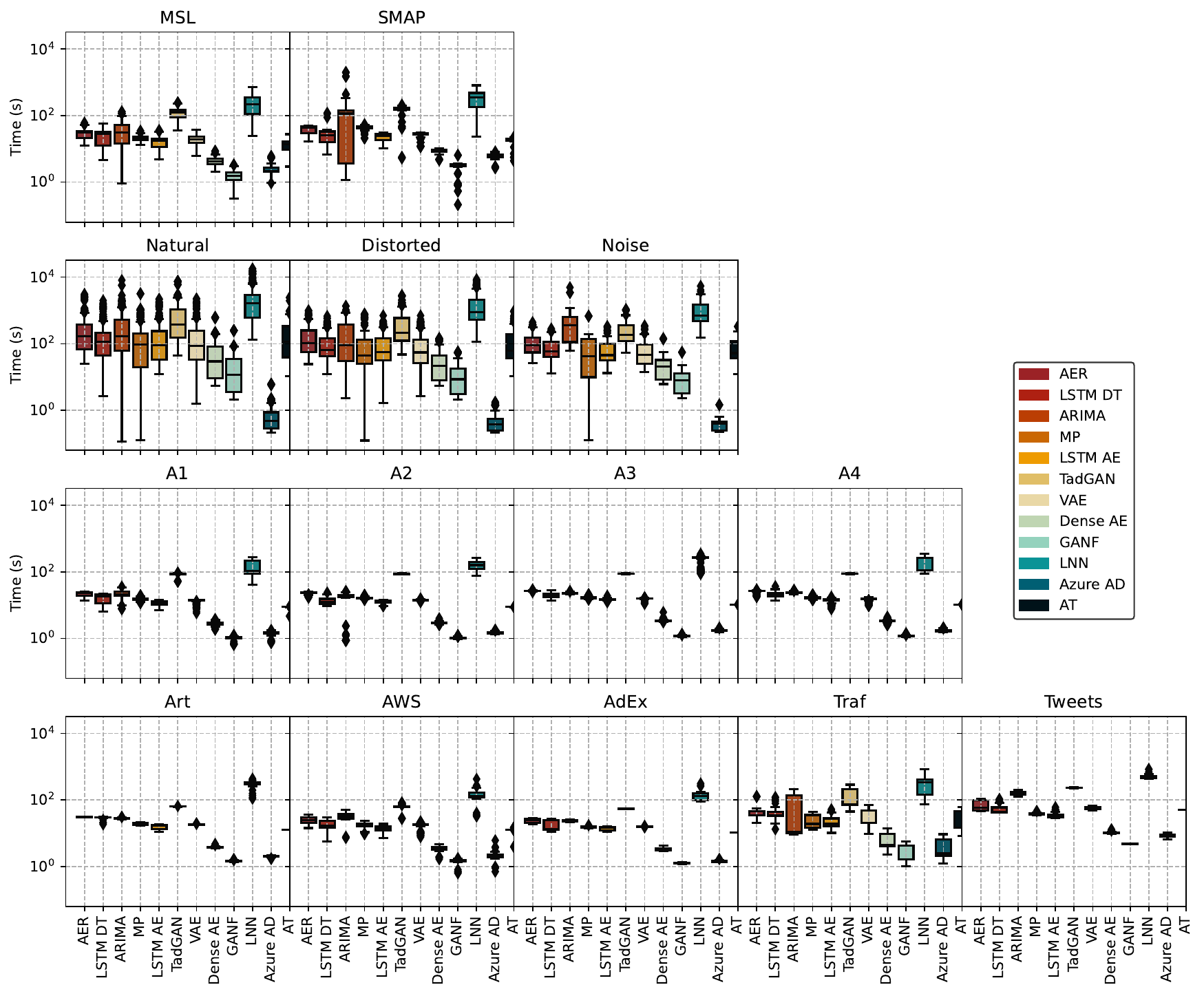}
    \caption{Runtime (in seconds) per dataset. Runtime is recorded as the wall time it takes to train a pipeline using \texttt{fit} and run inference using \texttt{detect}. The most time consuming pipeline is \texttt{TadGAN}.}
    \label{fig:runtime-boxplot}
\end{figure}

\subsection{Benchmark Results}
Figure~\ref{fig:f1-boxplot-dataset} illustrates the F1 score obtained per dataset.
Observing the average values per dataset, \texttt{AER} seems to score the highest on most datasets.
However, other pipelines such as \texttt{LSTM DT} are comparable or outperform \texttt{AER} in certain cases.
Each pipeline has its strengths and the performance varies from one dataset to another.

Pipeline scalability is an important aspect to address for many \user{s}.
The reported wall time of each pipeline per dataset is shown in Figure~\ref{fig:runtime-boxplot}.
\texttt{TadGAN} takes minutes to run while other pipelines seem to finish in several seconds.
The fastest pipelines are \texttt{GANF} and \texttt{Azure AD}.
\texttt{Azure AD} is an inference only pipeline, and \texttt{GANF} is fast to train.

% \begin{table}[htb]
%     \centering
%     \caption{Leaderboard showing the number of wins of a particular pipeline compared with classic pipeline ARIMA in terms of F1 Score on 12 datasets.}
%     \input{tables/leaderboard}
%     \label{tab:leaderboard}
% \end{table}

\subsection{Leaderboard}
% With every release, we present a leaderboard similar to Table~\ref{tab:leaderboard}.
% It depicts the number of times each pipeline outperformed \texttt{ARIMA} in F1 Score (maximum \# datasets).
% Specifically, Table~\ref{tab:leaderboard} is generated from benchmark version \texttt{0.5.2}.
% It provides an overall view of how deep learning models perform compared to a classical method such as \texttt{ARIMA}.
% \texttt{AER} fluctuates between 11 and 12, beating \texttt{ARIMA} on almost every dataset, with the exception of AdEx dataset.

Table~\ref{tab:ranks} shows the rank of each pipeline in 5 different benchmark runs.
The rank is calculated from the order of the leaderboard (as shown in Table~\ref{tab:leaderboard}).
If two pipelines have the same number of wins, the average F1 score is used as a tie-breaker.
\begin{table}[ht]
    \centering
    \caption{Rank of pipelines in five independent runs.}
    \label{tab:ranks}
    \begin{tabular}{lccccc}
    \toprule
    &       \multicolumn{5}{c}{\textbf{Run}}\\
    \cmidrule(lr){2-6}
    \textbf{Pipeline}               &  \#1 &  \#2 &  \#3 &  \#4 &  \#5 \\
    \midrule
    \texttt{AER}            &     1 &     1 &     1 &     1 &     1 \\
    \texttt{LSTM DT}        &     3 &     2 &     2 &     2 &     2 \\
    \texttt{LSTM AE}        &     2 &     5 &     3 &     4 &     4 \\
    \texttt{TadGAN}         &     6 &     7 &     5 &     5 &     6 \\
    \texttt{VAE}            &     4 &     3 &     4 &     7 &     5 \\
    \texttt{Dense AE}       &     7 &     4 &     6 &     6 &     7 \\
    \texttt{LNN}            &     5 &     6 &     7 &     3 &     3 \\
    \texttt{GANF}           &     8 &     8 &     8 &     8 &     8 \\
    \texttt{MP}             &     9 &     9 &     9 &     9 &     9 \\
    \texttt{AT}             &    10 &    10 &    10 &    10 &    10 \\
    \texttt{Azure AD}       &    11 &    11 &    11 &    11 &    11 \\
    \bottomrule
    \end{tabular}
\end{table}

\subsection{Computational Cost Across Releases}

\begin{figure}[htb]
    \centering
    \includegraphics[width=.7\linewidth]{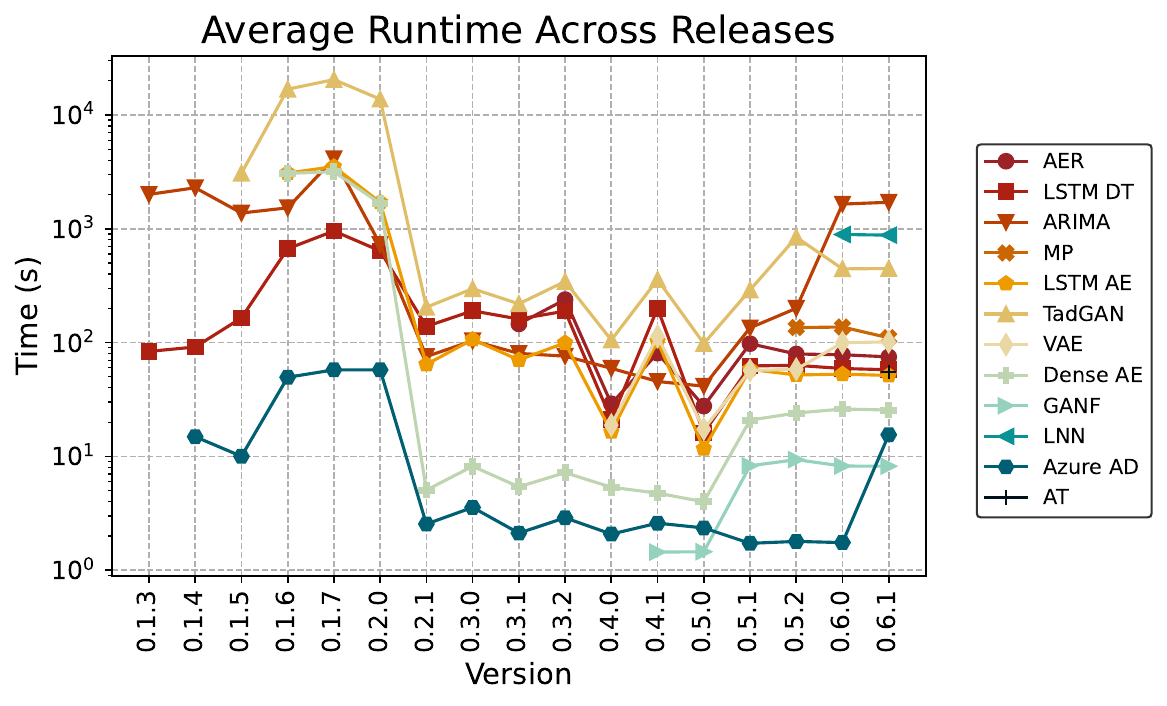}
    \caption{Average runtimes (in seconds) across Orion versions. Deep learning pipelines with LSTM layers are more sporadic across releases due to the availability of GPU. Pipelines such as \texttt{Azure AD} are black box pipelines that run inference alone making it fast. In version \texttt{0.2.1}, we migrated our benchmark to MIT Supercloud.}
    \label{fig:runtime-version}
\end{figure}

In addition to quality stability shown in Figure~\ref{fig:version}, we can monitor the runtime execution of the benchmark.
We illustrate the average runtime for each pipeline across 15 releases in Figure~\ref{fig:runtime-version}.
There is a clear improvement in average runtime in release \texttt{0.2.1}.
This increase in speed traces back to an internal change of the API's code.
More specifically, pipelines builds were adjusted to only build once to reduce overhead during the \texttt{fit} and \texttt{detect} process. Looking back at the development plan, this is reflected in \href{https://github.com/sintel-dev/Orion/issues/261}{Issue \#261} where we see exact alterations made to the code.

Moreover, in version \texttt{0.4.0}, the package migrated to \texttt{TensorFlow 2.0} which consequently made the pipelines faster in GPU mode. However, in version \texttt{0.4.1} the pipelines were executed without GPU, which is evident by the slight increase in runtime. 
% Generally, pipelines with LSTM and convolutional layers benefit greatly from GPUs.

\subsection{Qualitative Performance Across Releases}
The detailed sheets of benchmark runs are stored directly in the repository: \url{https://github.com/sintel-dev/Orion/tree/master/benchmark/results}.
The following tables report the F1 score, precision, and recall metrics for each pipeline across all releases as of today.

\begin{itemize}
    \item version \texttt{0.6.1}, pipelines 12, datasets 12, release date: October 4\textsuperscript{th} 2024.
    \item version \texttt{0.6.0}, pipelines 11, datasets 12, release date: February 13\textsuperscript{th} 2024.
    \item version \texttt{0.5.2}, pipelines 10, datasets 12, release date: October 19\textsuperscript{th} 2023.
    \item version \texttt{0.5.1}, pipelines 9, datasets 12, release date: August 16\textsuperscript{th} 2023.
    \item version \texttt{0.5.0}, pipelines 9, datasets 11, release date: May 23\textsuperscript{rd} 2023.
    \item version \texttt{0.4.1}, pipelines 9, datasets 11, release date: January 31\textsuperscript{st} 2023.
    \item version \texttt{0.4.0}, pipelines 8, datasets 11, release date: November 10\textsuperscript{th} 2022.
    \item version \texttt{0.3.2}, pipelines 7, datasets 11, release date: July 4\textsuperscript{th} 2022.
    \item version \texttt{0.3.1}, pipelines 7, datasets 11, release date: April 26\textsuperscript{th} 2022.
    \item version \texttt{0.3.0}, pipelines 6, datasets 11, release date: March 31\textsuperscript{st} 2022.
    \item version \texttt{0.2.1}, pipelines 6, datasets 11, release date: February 18\textsuperscript{th} 2022.
    \item version \texttt{0.2.0}, pipelines 6, datasets 11, release date: October 11\textsuperscript{th} 2021.
    \item version \texttt{0.1.7}, pipelines 6, datasets 11, release date: May 4\textsuperscript{th} 2021.
    \item version \texttt{0.1.6}, pipelines 6, datasets 11, release date: March 8\textsuperscript{th} 2021.
    \item version \texttt{0.1.5}, pipelines 4, datasets 11, release date: December 25\textsuperscript{th} 2020.
    \item version \texttt{0.1.4}, pipelines 3, datasets 11, release date: October 16\textsuperscript{th} 2020.
    \item version \texttt{0.1.3}, pipelines 2, datasets 11, release date: September 29\textsuperscript{th} 2020.
\end{itemize}

\newcommand{\aer}{\texttt{AER}}{}
\newcommand{\lstmdt}{\texttt{LSTM DT}}{}
\newcommand{\arima}{\texttt{ARIMA}}{}
\newcommand{\pmatrixp}{\texttt{MP}}{}
\newcommand{\lstmae}{\texttt{LSTM AE}}{}
\newcommand{\tadgan}{\texttt{TadGAN}}{}
\newcommand{\vae}{\texttt{VAE}}{}
\newcommand{\dense}{\texttt{Dense AE}}{}
\newcommand{\ganf}{\texttt{GANF}}{}
\newcommand{\lnn}{\texttt{LNN}}{}
\newcommand{\azure}{\texttt{Azure AD}}{}
\newcommand{\trans}{\texttt{AT}}{}  

%%%%%%%%%%%%%%%%%%%%%%%
%%%%%%% Overall %%%%%%%
%%%%%%%%%%%%%%%%%%%%%%%
% \begin{adjustbox}{angle=90, caption={Benchmark Results}, float=table}
% \begin{table*}[!t]
\begin{sidewaystable}
% \caption{Benchmark Results}
\label{tab:bench_f1}
\resizebox{\textwidth}{!}{%
\begin{tabular}{lcccccccccccccc}
\toprule
& \multicolumn{2}{c}{NASA} & \multicolumn{3}{c}{UCR} & \multicolumn{4}{c}{Yahoo S5} & \multicolumn{5}{c}{NAB}  \\
\cmidrule(r){2-3}\cmidrule(r){4-6}\cmidrule(r){7-10}\cmidrule(r){11-15}
\textbf{Pipeline} & MSL & SMAP & Natural & Distorted & Noise & A1 & A2 & A3 & A4 & Art & AWS & AdEx & Traf & Tweets\\
\midrule
\multicolumn{12}{c}{\textbf{F1 Score}}\\
\midrule
\aer                &  0.605$\pm$0.02 &  0.792$\pm$0.02 &  0.427$\pm$0.01 &  0.547$\pm$0.01 &  0.546$\pm$0.05 &  0.784$\pm$0.01 &  0.984$\pm$0.00 &  0.889$\pm$0.01 &  0.712$\pm$0.01 
                    &  0.682$\pm$0.16 &  0.737$\pm$0.02 &  0.708$\pm$0.04 &  0.691$\pm$0.03 &  0.565$\pm$0.02 \\
\lstmdt             &  0.483$\pm$0.01 &  0.713$\pm$0.02 &  0.386$\pm$0.01 &  0.411$\pm$0.02 &  0.460$\pm$0.03 &  0.736$\pm$0.02 &  0.978$\pm$0.00 &  0.738$\pm$0.01 &  0.649$\pm$0.01 
                    &  0.404$\pm$0.03 &  0.495$\pm$0.03 &  0.734$\pm$0.02 &  0.634$\pm$0.04 &  0.556$\pm$0.01 \\
\arima              &  0.435$\pm$0.00 &  0.324$\pm$0.00 &  0.077$\pm$0.00 &  0.092$\pm$0.00 &  0.176$\pm$0.00 &  0.744$\pm$0.00 &  0.816$\pm$0.00 &  0.782$\pm$0.00 &  0.684$\pm$0.00 
                    &  0.429$\pm$0.00 &  0.472$\pm$0.00 &  0.727$\pm$0.00 &  0.429$\pm$0.00 &  0.513$\pm$0.00 \\
\pmatrixp           &  0.474$\pm$0.00 &  0.423$\pm$0.00 &  0.259$\pm$0.00 &  0.086$\pm$0.00 &  0.037$\pm$0.00 &  0.507$\pm$0.00 &  0.897$\pm$0.00 &  0.793$\pm$0.00 &  0.825$\pm$0.00 
                    &  0.571$\pm$0.00 &  0.440$\pm$0.00 &  0.692$\pm$0.00 &  0.310$\pm$0.00 &  0.343$\pm$0.00 \\
\lstmae             &  0.493$\pm$0.05 &  0.672$\pm$0.03 &  0.281$\pm$0.01 &  0.375$\pm$0.01 &  0.419$\pm$0.04 &  0.595$\pm$0.01 &  0.871$\pm$0.02 &  0.444$\pm$0.01 &  0.238$\pm$0.01 
                    &  0.594$\pm$0.07 &  0.754$\pm$0.02 &  0.665$\pm$0.06 &  0.490$\pm$0.02 &  0.513$\pm$0.02 \\
\bottomrule
\end{tabular}}
% \end{table*}
\end{sidewaystable}

%%%%%%%%%%%%%%%%%%%%%%%
%%%%%%%% 0.6.1 %%%%%%%%
%%%%%%%%%%%%%%%%%%%%%%%
\begin{table*}[!t]
\caption{Benchmark Summary Results Version \texttt{0.6.1}}
\resizebox{\linewidth}{!}{%
\begin{tabular}{lcccccccccccc}
\toprule
& \multicolumn{2}{c}{NASA} & UCR & \multicolumn{4}{c}{Yahoo S5} & \multicolumn{5}{c}{NAB}  \\
\cmidrule(r){2-3}\cmidrule(r){4-4}\cmidrule(r){5-8}\cmidrule(r){9-13}
\textbf{Pipeline} & MSL & SMAP & UCR & A1 & A2 & A3 & A4 & Art & AWS & AdEx & Traf & Tweets\\
\midrule
\multicolumn{13}{c}{\textbf{F1 Score}}\\
\midrule
\aer        & 0.533 & 0.781 & 0.489 & 0.784 & 0.987 & 0.878 & 0.712 & 0.714 & 0.727 & 0.690 & 0.703 & 0.562 \\
\lstmdt     & 0.466 & 0.694 & 0.390 & 0.735 & 0.980 & 0.743 & 0.637 & 0.400 & 0.507 & 0.714 & 0.585 & 0.603 \\
\arima      & 0.525 & 0.411 & 0.153 & 0.728 & 0.856 & 0.797 & 0.686 & 0.308 & 0.382 & 0.727 & 0.467 & 0.514 \\
\pmatrixp   & 0.474 & 0.423 & 0.051 & 0.507 & 0.897 & 0.793 & 0.825 & 0.571 & 0.440 & 0.692 & 0.305 & 0.343 \\
\lstmae     & 0.462 & 0.662 & 0.330 & 0.600 & 0.864 & 0.444 & 0.247 & 0.667 & 0.737 & 0.500 & 0.467 & 0.508 \\
\bottomrule
\end{tabular}}
\end{table*}

%%%%%%%%%%%%%%%%%%%%%%%
%%%%%%%% 0.6.0 %%%%%%%%
%%%%%%%%%%%%%%%%%%%%%%%
\begin{table*}[!t]
\caption{Benchmark Summary Results Version \texttt{0.6.0}}
\resizebox{\linewidth}{!}{%
\begin{tabular}{lcccccccccccc}
\toprule
& \multicolumn{2}{c}{NASA} & UCR & \multicolumn{4}{c}{Yahoo S5} & \multicolumn{5}{c}{NAB}  \\
\cmidrule(r){2-3}\cmidrule(r){4-4}\cmidrule(r){5-8}\cmidrule(r){9-13}
\textbf{Pipeline} & MSL & SMAP & UCR & A1 & A2 & A3 & A4 & Art & AWS & AdEx & Traf & Tweets\\
\midrule
\multicolumn{13}{c}{\textbf{F1 Score}}\\
\midrule
\aer        &  0.587 &  0.819 &  0.476 &  0.799 &  0.987 &  0.892 &  0.709 &  0.714 &  0.741 &  0.690 &  0.703 &  0.638 \\
\lstmdt     &  0.471 &  0.726 &  0.393 &  0.728 &  0.985 &  0.744 &  0.646 &  0.400 &  0.468 &  0.786 &  0.585 &  0.603 \\
\arima      &  0.525 &  0.411 &  0.153 &  0.728 &  0.856 &  0.797 &  0.686 &  0.308 &  0.382 &  0.727 &  0.467 &  0.514 \\
\pmatrixp   &  0.474 &  0.423 &  0.051 &  0.507 &  0.897 &  0.793 &  0.825 &  0.571 &  0.440 &  0.692 &  0.305 &  0.343 \\
\lstmae     &  0.545 &  0.662 &  0.327 &  0.595 &  0.867 &  0.466 &  0.239 &  0.667 &  0.741 &  0.500 &  0.500 &  0.475 \\
\bottomrule
\end{tabular}}
\end{table*}

%%%%%%%%%%%%%%%%%%%%%%%
%%%%%%%% 0.5.2 %%%%%%%%
%%%%%%%%%%%%%%%%%%%%%%%
\begin{table*}[!t]
\caption{Benchmark Summary Results Version \texttt{0.5.2}}
\resizebox{\linewidth}{!}{%
\begin{tabular}{lcccccccccccc}
\toprule
& \multicolumn{2}{c}{NASA} & UCR & \multicolumn{4}{c}{Yahoo S5} & \multicolumn{5}{c}{NAB}  \\
\cmidrule(r){2-3}\cmidrule(r){4-4}\cmidrule(r){5-8}\cmidrule(r){9-13}
\textbf{Pipeline} & MSL & SMAP & UCR & A1 & A2 & A3 & A4 & Art & AWS & AdEx & Traf & Tweets\\
\midrule
\multicolumn{13}{c}{\textbf{F1 Score}}\\
\midrule
\aer        &  0.587 &  0.775 &  0.474 &  0.780 &  0.987 &  0.869 &  0.686 &  0.769 &  0.750 &  0.733 &  0.611 &  0.581 \\
\lstmdt     &  0.485 &  0.707 &  0.417 &  0.724 &  0.987 &  0.744 &  0.644 &  0.400 &  0.494 &  0.759 &  0.667 &  0.600 \\
\arima      &  0.435 &  0.326 &  0.090 &  0.744 &  0.816 &  0.782 &  0.684 &  0.429 &  0.472 &  0.727 &  0.429 &  0.513 \\
\pmatrixp   &  0.474 &  0.423 &  0.051 &  0.507 &  0.897 &  0.793 &  0.825 &  0.571 &  0.440 &  0.692 &  0.305 &  0.343 \\
\lstmae     &  0.479 &  0.662 &  0.332 &  0.619 &  0.874 &  0.460 &  0.227 &  0.667 &  0.750 &  0.615 &  0.471 &  0.533 \\
\bottomrule
\end{tabular}}
\end{table*}

%%%%%%%%%%%%%%%%%%%%%%%
%%%%%%%% 0.5.1 %%%%%%%%
%%%%%%%%%%%%%%%%%%%%%%%
\begin{table*}[!t]
\caption{Benchmark Summary Results Version \texttt{0.5.1}}
\resizebox{\linewidth}{!}{%
\begin{tabular}{lcccccccccccc}
\toprule
& \multicolumn{2}{c}{NASA} & UCR & \multicolumn{4}{c}{Yahoo S5} & \multicolumn{5}{c}{NAB}  \\
\cmidrule(r){2-3}\cmidrule(r){4-4}\cmidrule(r){5-8}\cmidrule(r){9-13}
\textbf{Pipeline} & MSL & SMAP & UCR & A1 & A2 & A3 & A4 & Art & AWS & AdEx & Traf & Tweets\\
\midrule
\multicolumn{13}{c}{\textbf{F1 Score}}\\
\midrule
\aer        &  0.575 &  0.803 &  0.482 &  0.799 &  0.987 &  0.898 &  0.712 &  0.500 &  0.750 &  0.667 &  0.703 &  0.571 \\
\lstmdt     &  0.471 &  0.730 &  0.393 &  0.743 &  0.980 &  0.734 &  0.639 &  0.400 &  0.494 &  0.710 &  0.632 &  0.560 \\
\arima      &  0.435 &  0.326 &  0.090 &  0.744 &  0.816 &  0.782 &  0.684 &  0.429 &  0.472 &  0.727 &  0.429 &  0.513 \\
\lstmae     &  0.533 &  0.658 &  0.311 &  0.593 &  0.852 &  0.452 &  0.252 &  0.545 &  0.737 &  0.667 &  0.500 &  0.542 \\
\tadgan     &  0.571 &  0.592 &  0.172 &  0.547 &  0.809 &  0.427 &  0.324 &  0.667 &  0.645 &  0.727 &  0.486 &  0.552 \\
\bottomrule
\end{tabular}}
\end{table*}

%%%%%%%%%%%%%%%%%%%%%%%
%%%%%%%% 0.5.0 %%%%%%%%
%%%%%%%%%%%%%%%%%%%%%%%
\begin{table*}[!t]
\caption{Benchmark Summary Results Version \texttt{0.5.0}}
\resizebox{\linewidth}{!}{%
\begin{tabular}{lccccccccccc}
\toprule
& \multicolumn{2}{c}{NASA} & \multicolumn{4}{c}{Yahoo S5} & \multicolumn{5}{c}{NAB}  \\
\cmidrule(r){2-3}\cmidrule(r){4-7}\cmidrule(r){8-12}
\textbf{Pipeline} & MSL & SMAP & A1 & A2 & A3 & A4 & Art & AWS & AdEx & Traf & Tweets\\
\midrule
\multicolumn{12}{c}{\textbf{F1 Score}}\\
\midrule
\aer        &  0.632 &  0.784 &  0.767 &  0.978 &  0.878 &  0.708 &  0.769 &  0.727 &  0.667 &  0.686 &  0.603 \\
\lstmdt     &  0.481 &  0.708 &  0.726 &  0.973 &  0.740 &  0.638 &  0.400 &  0.474 &  0.759 &  0.649 &  0.560 \\
\arima      &  0.435 &  0.324 &  0.744 &  0.816 &  0.782 &  0.684 &  0.429 &  0.472 &  0.727 &  0.429 &  0.513 \\
\lstmae     &  0.514 &  0.686 &  0.600 &  0.888 &  0.443 &  0.227 &  0.667 &  0.800 &  0.522 &  0.500 &  0.483 \\
\tadgan     &  0.482 &  0.573 &  0.612 &  0.818 &  0.371 &  0.327 &  0.615 &  0.645 &  0.538 &  0.512 &  0.551 \\
\bottomrule
\end{tabular}}
\end{table*}

%%%%%%%%%%%%%%%%%%%%%%%
%%%%%%%% 0.4.1 %%%%%%%%
%%%%%%%%%%%%%%%%%%%%%%%
\begin{table*}[!t]
\caption{Benchmark Summary Results Version \texttt{0.4.1}}
\resizebox{\linewidth}{!}{%
\begin{tabular}{lccccccccccc}
\toprule
& \multicolumn{2}{c}{NASA} & \multicolumn{4}{c}{Yahoo S5} & \multicolumn{5}{c}{NAB}  \\
\cmidrule(r){2-3}\cmidrule(r){4-7}\cmidrule(r){8-12}
\textbf{Pipeline} & MSL & SMAP & A1 & A2 & A3 & A4 & Art & AWS & AdEx & Traf & Tweets\\
\midrule
\multicolumn{12}{c}{\textbf{F1 Score}}\\
\midrule
\aer        &  0.583 &  0.778 &  0.787 &  0.978 &  0.895 &  0.691 &  0.750 &  0.690 &  0.733 &  0.632 &  0.606 \\
\lstmdt     &  0.457 &  0.707 &  0.743 &  0.980 &  0.748 &  0.651 &  0.421 &  0.474 &  0.733 &  0.649 &  0.571 \\
\arima      &  0.435 &  0.324 &  0.744 &  0.816 &  0.782 &  0.684 &  0.429 &  0.472 &  0.727 &  0.429 &  0.513 \\
\lstmae     &  0.493 &  0.662 &  0.588 &  0.885 &  0.446 &  0.237 &  0.667 &  0.712 &  0.615 &  0.552 &  0.492 \\
\tadgan     &  0.543 &  0.620 &  0.558 &  0.828 &  0.428 &  0.321 &  0.571 &  0.585 &  0.583 &  0.516 &  0.559 \\
\bottomrule
\end{tabular}}
\end{table*}

%%%%%%%%%%%%%%%%%%%%%%%
%%%%%%%% 0.4.0 %%%%%%%%
%%%%%%%%%%%%%%%%%%%%%%%
\begin{table*}[!t]
\caption{Benchmark Summary Results Version \texttt{0.4.0}}
\resizebox{\linewidth}{!}{%
\begin{tabular}{lccccccccccc}
\toprule
& \multicolumn{2}{c}{NASA} & \multicolumn{4}{c}{Yahoo S5} & \multicolumn{5}{c}{NAB}  \\
\cmidrule(r){2-3}\cmidrule(r){4-7}\cmidrule(r){8-12}
\textbf{Pipeline} & MSL & SMAP & A1 & A2 & A3 & A4 & Art & AWS & AdEx & Traf & Tweets\\
\midrule
\multicolumn{12}{c}{\textbf{F1 Score}}\\
\midrule
\aer        &  0.579 &  0.770 &  0.793 &  0.978 &  0.888 &  0.721 &  0.800 &  0.727 &  0.690 &  0.667 &  0.571 \\
\lstmdt     &  0.472 &  0.717 &  0.744 &  0.987 &  0.735 &  0.652 &  0.400 &  0.545 &  0.759 &  0.585 &  0.579 \\
\arima      &  0.435 &  0.324 &  0.744 &  0.816 &  0.782 &  0.684 &  0.429 &  0.472 &  0.727 &  0.429 &  0.513 \\
\lstmae     &  0.548 &  0.681 &  0.611 &  0.877 &  0.456 &  0.233 &  0.545 &  0.764 &  0.667 &  0.452 &  0.500 \\
\tadgan     &  0.558 &  0.610 &  0.568 &  0.824 &  0.427 &  0.320 &  0.471 &  0.656 &  0.720 &  0.556 &  0.559 \\
\bottomrule
\end{tabular}}
\end{table*}

%%%%%%%%%%%%%%%%%%%%%%%
%%%%%%%% 0.3.2 %%%%%%%%
%%%%%%%%%%%%%%%%%%%%%%%
\begin{table*}[!t]
\caption{Benchmark Summary Results Version \texttt{0.3.2}}
\resizebox{\linewidth}{!}{%
\begin{tabular}{lccccccccccc}
\toprule
& \multicolumn{2}{c}{NASA} & \multicolumn{4}{c}{Yahoo S5} & \multicolumn{5}{c}{NAB}  \\
\cmidrule(r){2-3}\cmidrule(r){4-7}\cmidrule(r){8-12}
\textbf{Pipeline} & MSL & SMAP & A1 & A2 & A3 & A4 & Art & AWS & AdEx & Traf & Tweets\\
\midrule
\multicolumn{12}{c}{\textbf{F1 Score}}\\
\midrule
\aer        &  0.600 &  0.785 &  0.745 &  0.985 &  0.881 &  0.721 &  0.471 &  0.727 &  0.733 &  0.600 &  0.562 \\
\lstmdt     &  0.500 &  0.680 &  0.741 &  0.978 &  0.734 &  0.633 &  0.400 &  0.481 &  0.786 &  0.611 &  0.603 \\
\arima      &  0.344 &  0.309 &  0.744 &  0.816 &  0.782 &  0.684 &  0.429 &  0.472 &  0.727 &  0.429 &  0.513 \\
\lstmae     &  0.493 &  0.715 &  0.620 &  0.863 &  0.447 &  0.241 &  0.667 &  0.712 &  0.609 &  0.516 &  0.516 \\
\tadgan     &  0.595 &  0.645 &  0.531 &  0.829 &  0.416 &  0.350 &  0.714 &  0.645 &  0.818 &  0.541 &  0.580 \\
\bottomrule
\end{tabular}}
\end{table*}

%%%%%%%%%%%%%%%%%%%%%%%
%%%%%%%% 0.3.1 %%%%%%%%
%%%%%%%%%%%%%%%%%%%%%%%
\begin{table*}[!t]
\caption{Benchmark Summary Results Version \texttt{0.3.1}}
\resizebox{\linewidth}{!}{%
\begin{tabular}{lccccccccccc}
\toprule
& \multicolumn{2}{c}{NASA} & \multicolumn{4}{c}{Yahoo S5} & \multicolumn{5}{c}{NAB}  \\
\cmidrule(r){2-3}\cmidrule(r){4-7}\cmidrule(r){8-12}
\textbf{Pipeline} & MSL & SMAP & A1 & A2 & A3 & A4 & Art & AWS & AdEx & Traf & Tweets\\
\midrule
\multicolumn{12}{c}{\textbf{F1 Score}}\\
\midrule
\aer        &  0.579 &  0.778 &  0.786 &  0.992 &  0.896 &  0.716 &  0.533 &  0.678 &  0.759 &  0.667 &  0.581 \\
\lstmdt     &  0.486 &  0.703 &  0.752 &  0.985 &  0.743 &  0.635 &  0.400 &  0.545 &  0.733 &  0.667 &  0.580 \\
\arima      &  0.344 &  0.309 &  0.744 &  0.816 &  0.782 &  0.684 &  0.429 &  0.472 &  0.727 &  0.429 &  0.513 \\
\lstmae     &  0.500 &  0.690 &  0.625 &  0.867 &  0.446 &  0.238 &  0.667 &  0.764 &  0.583 &  0.500 &  0.500 \\
\tadgan     &  0.512 &  0.658 &  0.566 &  0.858 &  0.422 &  0.331 &  0.714 &  0.625 &  0.750 &  0.588 &  0.559 \\
\bottomrule
\end{tabular}}
\end{table*}

%%%%%%%%%%%%%%%%%%%%%%%
%%%%%%%% 0.3.0 %%%%%%%%
%%%%%%%%%%%%%%%%%%%%%%%
\begin{table*}[!t]
\caption{Benchmark Summary Results Version \texttt{0.3.0}}
\resizebox{\linewidth}{!}{%
\begin{tabular}{lccccccccccc}
\toprule
& \multicolumn{2}{c}{NASA} & \multicolumn{4}{c}{Yahoo S5} & \multicolumn{5}{c}{NAB}  \\
\cmidrule(r){2-3}\cmidrule(r){4-7}\cmidrule(r){8-12}
\textbf{Pipeline} & MSL & SMAP & A1 & A2 & A3 & A4 & Art & AWS & AdEx & Traf & Tweets\\
\midrule
\multicolumn{12}{c}{\textbf{F1 Score}}\\
\midrule
\lstmdt     &  0.476 &  0.741 &  0.739 &  0.990 &  0.753 &  0.644 &  0.400 &  0.537 &  0.714 &  0.703 &  0.556 \\
\arima      &  0.344 &  0.309 &  0.744 &  0.816 &  0.782 &  0.684 &  0.429 &  0.472 &  0.727 &  0.429 &  0.513 \\
\lstmae     &  0.500 &  0.658 &  0.584 &  0.877 &  0.444 &  0.262 &  0.667 &  0.724 &  0.667 &  0.471 &  0.475 \\
\tadgan     &  0.575 &  0.659 &  0.564 &  0.853 &  0.439 &  0.370 &  0.615 &  0.656 &  0.640 &  0.541 &  0.559 \\
\dense      &  0.523 &  0.661 &  0.665 &  0.891 &  0.072 &  0.109 &  0.400 &  0.764 &  0.571 &  0.552 &  0.517 \\
\bottomrule
\end{tabular}}
\end{table*}

%%%%%%%%%%%%%%%%%%%%%%%
%%%%%%%% 0.2.1 %%%%%%%%
%%%%%%%%%%%%%%%%%%%%%%%
\begin{table*}[!t]
\caption{Benchmark Summary Results Version \texttt{0.2.1}}
\resizebox{\linewidth}{!}{%
\begin{tabular}{lccccccccccc}
\toprule
& \multicolumn{2}{c}{NASA} & \multicolumn{4}{c}{Yahoo S5} & \multicolumn{5}{c}{NAB}  \\
\cmidrule(r){2-3}\cmidrule(r){4-7}\cmidrule(r){8-12}
\textbf{Pipeline} & MSL & SMAP & A1 & A2 & A3 & A4 & Art & AWS & AdEx & Traf & Tweets\\
\midrule
\multicolumn{12}{c}{\textbf{F1 Score}}\\
\midrule
\lstmdt     &  0.460 &  0.703 &  0.752 &  0.980 &  0.733 &  0.643 &  0.400 &  0.481 &  0.643 &  0.684 &  0.568 \\
\arima      &  0.344 &  0.309 &  0.744 &  0.816 &  0.782 &  0.684 &  0.429 &  0.472 &  0.727 &  0.429 &  0.513 \\
\lstmae     &  0.480 &  0.690 &  0.600 &  0.870 &  0.436 &  0.243 &  0.545 &  0.750 &  0.615 &  0.571 &  0.525 \\
\tadgan     &  0.558 &  0.650 &  0.559 &  0.890 &  0.412 &  0.374 &  0.500 &  0.677 &  0.692 &  0.500 &  0.567 \\
\dense      &  0.529 &  0.661 &  0.652 &  0.899 &  0.087 &  0.092 &  0.545 &  0.741 &  0.632 &  0.552 &  0.517 \\
\bottomrule
\end{tabular}}
\end{table*}

%%%%%%%%%%%%%%%%%%%%%%%
%%%%%%%% 0.2.0 %%%%%%%%
%%%%%%%%%%%%%%%%%%%%%%%
\begin{table*}[!t]
\caption{Benchmark Summary Results Version \texttt{0.2.0}}
\resizebox{\linewidth}{!}{%
\begin{tabular}{lccccccccccc}
\toprule
& \multicolumn{2}{c}{NASA} & \multicolumn{4}{c}{Yahoo S5} & \multicolumn{5}{c}{NAB}  \\
\cmidrule(r){2-3}\cmidrule(r){4-7}\cmidrule(r){8-12}
\textbf{Pipeline} & MSL & SMAP & A1 & A2 & A3 & A4 & Art & AWS & AdEx & Traf & Tweets\\
\midrule
\multicolumn{12}{c}{\textbf{F1 Score}}\\
\midrule
\lstmdt     &  0.447 &  0.671 &  0.724 &  0.975 &  0.751 &  0.637 &  0.400 &  0.488 &  0.733 &  0.619 &  0.535 \\
\arima      &  0.435 &  0.326 &  0.744 &  0.816 &  0.782 &  0.684 &  0.429 &  0.472 &  0.727 &  0.429 &  0.513 \\
\lstmae     &  0.500 &  0.667 &  0.593 &  0.859 &  0.422 &  0.223 &  0.667 &  0.724 &  0.720 &  0.552 &  0.540 \\
\tadgan     &  0.465 &  0.646 &  0.549 &  0.843 &  0.492 &  0.388 &  0.667 &  0.623 &  0.741 &  0.476 &  0.523 \\
\dense      &  0.563 &  0.710 &  0.656 &  0.889 &  0.084 &  0.094 &  0.545 &  0.800 &  0.632 &  0.500 &  0.517 \\
\bottomrule
\end{tabular}}
\end{table*}

%%%%%%%%%%%%%%%%%%%%%%%
%%%%%%%% 0.1.7 %%%%%%%%
%%%%%%%%%%%%%%%%%%%%%%%
\begin{table*}[!t]
\caption{Benchmark Summary Results Version \texttt{0.1.7}}
\resizebox{\linewidth}{!}{%
\begin{tabular}{lccccccccccc}
\toprule
& \multicolumn{2}{c}{NASA} & \multicolumn{4}{c}{Yahoo S5} & \multicolumn{5}{c}{NAB}  \\
\cmidrule(r){2-3}\cmidrule(r){4-7}\cmidrule(r){8-12}
\textbf{Pipeline} & MSL & SMAP & A1 & A2 & A3 & A4 & Art & AWS & AdEx & Traf & Tweets\\
\midrule
\multicolumn{12}{c}{\textbf{F1 Score}}\\
\midrule
\lstmdt     &  0.466 &  0.689 &  0.739 &  0.975 &  0.746 &  0.645 &  0.400 &  0.500 &  0.759 &  0.585 &  0.551 \\
\arima      &  0.344 &  0.309 &  0.744 &  0.816 &  0.782 &  0.684 &  0.429 &  0.472 &  0.727 &  0.429 &  0.513 \\
\lstmae     &  0.507 &  0.667 &  0.601 &  0.880 &  0.445 &  0.231 &  0.667 &  0.712 &  0.667 &  0.516 &  0.508 \\
\tadgan     &  0.517 &  0.634 &  0.551 &  0.841 &  0.484 &  0.376 &  0.571 &  0.689 &  0.769 &  0.563 &  0.559 \\
\dense      &  0.507 &  0.693 &  0.665 &  0.904 &  0.078 &  0.090 &  0.545 &  0.786 &  0.600 &  0.581 &  0.533 \\
\bottomrule
\end{tabular}}
\end{table*}

%%%%%%%%%%%%%%%%%%%%%%%
%%%%%%%% 0.1.6 %%%%%%%%
%%%%%%%%%%%%%%%%%%%%%%%
\begin{table*}[!t]
\caption{Benchmark Summary Results Version \texttt{0.1.6}}
\resizebox{\linewidth}{!}{%
\begin{tabular}{lccccccccccc}
\toprule
& \multicolumn{2}{c}{NASA} & \multicolumn{4}{c}{Yahoo S5} & \multicolumn{5}{c}{NAB}  \\
\cmidrule(r){2-3}\cmidrule(r){4-7}\cmidrule(r){8-12}
\textbf{Pipeline} & MSL & SMAP & A1 & A2 & A3 & A4 & Art & AWS & AdEx & Traf & Tweets\\
\midrule
\multicolumn{12}{c}{\textbf{F1 Score}}\\
\midrule
\lstmdt     &  0.480 &  0.718 &  0.747 &  0.975 &  0.739 &  0.649 &  0.400 &  0.474 &  0.741 &  0.684 &  0.583 \\
\arima      &  0.344 &  0.309 &  0.744 &  0.816 &  0.782 &  0.684 &  0.429 &  0.472 &  0.727 &  0.429 &  0.513 \\
\lstmae     &  0.474 &  0.667 &  0.593 &  0.865 &  0.438 &  0.276 &  0.667 &  0.764 &  0.615 &  0.452 &  0.552 \\
\tadgan     &  0.529 &  0.654 &  0.555 &  0.822 &  0.487 &  0.377 &  0.714 &  0.645 &  0.741 &  0.486 &  0.567 \\
\dense      &  0.515 &  0.667 &  0.648 &  0.897 &  0.080 &  0.091 &  0.545 &  0.786 &  0.600 &  0.581 &  0.517 \\
\bottomrule
\end{tabular}}
\end{table*}

%%%%%%%%%%%%%%%%%%%%%%%
%%%%%%%% 0.1.5 %%%%%%%%
%%%%%%%%%%%%%%%%%%%%%%%
\begin{table*}[!t]
\caption{Benchmark Summary Results Version \texttt{0.1.5}}
\resizebox{\linewidth}{!}{%
\begin{tabular}{lccccccccccc}
\toprule
& \multicolumn{2}{c}{NASA} & \multicolumn{4}{c}{Yahoo S5} & \multicolumn{5}{c}{NAB}  \\
\cmidrule(r){2-3}\cmidrule(r){4-7}\cmidrule(r){8-12}
\textbf{Pipeline} & MSL & SMAP & A1 & A2 & A3 & A4 & Art & AWS & AdEx & Traf & Tweets\\
\midrule
\multicolumn{12}{c}{\textbf{F1 Score}}\\
\midrule
\lstmdt     &  0.532 &  0.704 &  0.735 &  0.980 &  0.743 &  0.653 &  0.400 &  0.462 &  0.467 &  0.615 &  0.548 \\
\arima      &  0.344 &  0.307 &  0.744 &  0.816 &  0.782 &  0.684 &  0.429 &  0.472 &  0.538 &  0.429 &  0.513 \\
\tadgan     &  0.575 &  0.644 &  0.626 &  0.700 &  0.494 &  0.381 &  0.714 &  0.677 &  0.800 &  0.450 &  0.592 \\
\azure      &  0.061 &  0.021 &  0.271 &  0.653 &  0.697 &  0.337 &  0.053 &  0.068 &  0.019 &  0.068 &  0.269 \\
\midrule
\multicolumn{12}{c}{\textbf{Precision}}\\
\midrule
\lstmdt     &  0.431 &  0.667 &  0.690 &  0.980 &  0.991 &  0.899 &  0.333 &  0.375 &  0.368 &  0.480 &  0.500 \\
\arima      &  0.393 &  0.300 &  0.684 &  0.772 &  0.998 &  0.955 &  0.375 &  0.405 &  0.467 &  0.429 &  0.444 \\
\tadgan     &  0.490 &  0.523 &  0.652 &  0.700 &  0.795 &  0.588 &  0.625 &  0.656 &  0.714 &  0.346 &  0.553 \\
\azure      &  0.032 &  0.011 &  0.163 &  0.484 &  0.541 &  0.212 &  0.027 &  0.036 &  0.009 &  0.035 &  0.161 \\
\midrule
\multicolumn{12}{c}{\textbf{Recall}}\\
\midrule
\lstmdt     &  0.694 &  0.746 &  0.787 &  0.980 &  0.594 &  0.513 &  0.500 &  0.600 &  0.636 &  0.857 &  0.606 \\
\arima      &  0.306 &  0.313 &  0.815 &  0.865 &  0.643 &  0.533 &  0.500 &  0.567 &  0.636 &  0.429 &  0.606 \\
\tadgan     &  0.694 &  0.836 &  0.601 &  0.700 &  0.359 &  0.281 &  0.833 &  0.700 &  0.909 &  0.643 &  0.636 \\
\azure      &  0.806 &  0.940 &  0.787 &  1.000 &  0.978 &  0.816 &  1.000 &  0.733 &  0.909 &  1.000 &  0.818 \\
\bottomrule
\end{tabular}}
\end{table*}

%%%%%%%%%%%%%%%%%%%%%%%
%%%%%%%% 0.1.4 %%%%%%%%
%%%%%%%%%%%%%%%%%%%%%%%
\begin{table*}[!t]
\caption{Benchmark Summary Results Version \texttt{0.1.4}}
\resizebox{\linewidth}{!}{%
\begin{tabular}{lccccccccccc}
\toprule
& \multicolumn{2}{c}{NASA} & \multicolumn{4}{c}{Yahoo S5} & \multicolumn{5}{c}{NAB}  \\
\cmidrule(r){2-3}\cmidrule(r){4-7}\cmidrule(r){8-12}
\textbf{Pipeline} & MSL & SMAP & A1 & A2 & A3 & A4 & Art & AWS & AdEx & Traf & Tweets\\
\midrule
\multicolumn{12}{c}{\textbf{F1 Score}}\\
\midrule
\lstmdt     &  0.468 &  0.708 &  0.738 &  0.978 &  0.728 &  0.634 &  0.400 &  0.468 &  0.437 &  0.667 &  0.564 \\
\arima      &  0.344 &  0.307 &  0.744 &  0.816 &  0.782 &  0.684 &  0.429 &  0.472 &  0.538 &  0.429 &  0.513 \\
\azure      &  0.061 &  0.021 &  0.277 &  0.653 &  0.692 &  0.333 &  0.053 &  0.068 &  0.019 &  0.068 &  0.269 \\
\midrule
\multicolumn{12}{c}{\textbf{Precision}}\\
\midrule
\lstmdt     &  0.379 &  0.662 &  0.679 &  0.970 &  0.984 &  0.890 &  0.333 &  0.383 &  0.333 &  0.545 &  0.489 \\
\arima      &  0.393 &  0.300 &  0.684 &  0.772 &  0.998 &  0.955 &  0.375 &  0.405 &  0.467 &  0.429 &  0.444 \\
\azure      &  0.032 &  0.011 &  0.184 &  0.484 &  0.537 &  0.216 &  0.027 &  0.036 &  0.009 &  0.035 &  0.161 \\
\midrule
\multicolumn{12}{c}{\textbf{Recall}}\\
\midrule
\lstmdt     &  0.611 &  0.761 &  0.809 &  0.985 &  0.577 &  0.492 &  0.500 &  0.600 &  0.636 &  0.857 &  0.667 \\
\arima      &  0.306 &  0.313 &  0.815 &  0.865 &  0.643 &  0.533 &  0.500 &  0.567 &  0.636 &  0.429 &  0.606 \\
\azure      &  0.806 &  0.940 &  0.562 &  1.000 &  0.972 &  0.729 &  1.000 &  0.733 &  0.909 &  1.000 &  0.818 \\
\bottomrule
\end{tabular}}
\end{table*}

%%%%%%%%%%%%%%%%%%%%%%%
%%%%%%%% 0.1.3 %%%%%%%%
%%%%%%%%%%%%%%%%%%%%%%%
\begin{table*}[!t]
\caption{Benchmark Summary Results Version \texttt{0.1.3}}
\resizebox{\linewidth}{!}{%
\begin{tabular}{lccccccccccc}
\toprule
& \multicolumn{2}{c}{NASA} & \multicolumn{4}{c}{Yahoo S5} & \multicolumn{5}{c}{NAB}  \\
\cmidrule(r){2-3}\cmidrule(r){4-7}\cmidrule(r){8-12}
\textbf{Pipeline} & MSL & SMAP & A1 & A2 & A3 & A4 & Art & AWS & AdEx & Traf & Tweets\\
\midrule
\multicolumn{12}{c}{\textbf{F1 Score}}\\
\midrule
\lstmdt     &  0.495 &  0.750 &  0.757 &  0.987 &  0.756 &  0.643 &  0.400 &  0.531 &  0.452 &  0.718 &  0.620 \\
\arima      &  0.489 &  0.424 &  0.753 &  0.856 &  0.783 &  0.693 &  0.429 &  0.576 &  0.538 &  0.545 &  0.513 \\
\midrule
\multicolumn{12}{c}{\textbf{Precision}}\\
\midrule
\lstmdt     &  0.364 &  0.680 &  0.721 &  0.975 &  0.986 &  0.890 &  0.333 &  0.472 &  0.350 &  0.560 &  0.579 \\
\arima      &  0.393 &  0.300 &  0.690 &  0.772 &  0.998 &  0.955 &  0.375 &  0.567 &  0.467 &  0.429 &  0.444 \\
\midrule
\multicolumn{12}{c}{\textbf{Recall}}\\
\midrule
\lstmdt     &  0.774 &  0.836 &  0.798 &  1.000 &  0.613 &  0.503 &  0.500 &  0.607 &  0.636 &  1.000 &  0.667 \\
\arima      &  0.647 &  0.724 &  0.829 &  0.961 &  0.644 &  0.543 &  0.500 &  0.586 &  0.636 &  0.750 &  0.606 \\
\bottomrule
\end{tabular}}
\end{table*}

%%%%%%%%%%%%%%%%%%%%%%%%%%%%%%%%%%%%%%%%%%%%%%%%%%%%%%%%%%%%%%%%%%%%%%%%%%%%%%%
%%%%%%%%%%%%%%%%%%%%%%%%%%%%%%%%%%%%%%%%%%%%%%%%%%%%%%%%%%%%%%%%%%%%%%%%%%%%%%%

\end{document}